%% file: main.tex
\pdfoutput=1

\documentclass[11pt]{article}

\usepackage[final]{acl}

\input{latex/preamble/packages}

\title{When More Metadata Misleads: Probing Field-Level Grounding in LLM-Generated Museum Descriptions}
\input{latex/preamble/authors}

\begin{document}
\sloppy
\maketitle

\begin{abstract}
Metadata-conditioned generation is often framed as a data-to-text task in which additional input fields should improve grounding. Museum description generation challenges this assumption because catalogue records are heterogeneous, partially visual, and unevenly documented across geographic regions. We study how metadata composition shapes the generation of artefact descriptions through a two-phase framework. Phase~1 analyses natural metadata variation in a large museum collection, where field count, field type, image availability, and \textit{Place} co-vary. Phase~2 introduces controlled field probes that hold object identity fixed while varying metadata quantity, field type, and image access. Our results show that metadata quantity alone does not explain generation behaviour. Field type determines whether models ground, omit, or infer information, while images can both support visual grounding and increase unsupported interpretation. More broadly, our findings show that metadata-conditioned data-to-text evaluation should assess field- and claim-level support, rather than relying only on aggregate input coverage or reference similarity.
\end{abstract}

\input{latex/section-final/01-introduction}
\input{latex/section-final/02-related-work}
\input{latex/section-final/03-experiemtal-setup}
\input{latex/section-final/04-results}
\input{latex/section-final/05-discussion}
\input{latex/section-final/06-conclusion}

\input{latex/section-final/07-limitations}
\section*{Acknowledgments}

This research is supported by the Dutch Science Foundation (NWO) project HAICu with project number NWA.1518.22.105. All content represents the opinion of the authors, which is not necessarily shared or endorsed by their respective employers and/or sponsors.

\section*{Limitations}
\label{sec:limitations}

We distinguish contextual and content fields, but this distinction may not be absolute: some fields can behave differently depending on object type, record format, and catalogue practice. For example, \textit{Materials} is usually grounded reliably when present, but can still trigger unsupported object-type or quality inferences. Future work should examine whether field categories should be defined dynamically based on both schema role and object-specific usage.

Our claim-level factuality is evaluated with an automatic nugget-verification pipeline and two crowdworkers rather than expert curator annotation. We validate the pipeline through span checks and agreement with NLI-based field signals, but automated verification may still miss subtle curatorial distinctions. Future work should include expert review, especially for visually plausible claims that are not recorded in the catalogue metadata.

Finally, expert references are selective descriptions rather than exhaustive records, which partly explains the divergence between reference similarity and field-level support. Our evaluation treats catalogue metadata as the authoritative source of support, which is appropriate for studying schema-faithful generation, but catalogue records may omit details visible in the image or known from external sources. Future work should examine how humans and domain experts balance descriptiveness, visual plausibility, and catalogue fidelity, and should distinguish catalogue-supported, image-supported, and externally supported claims rather than treating all unsupported content as equivalent.

\section*{Ethical Considerations}
\label{sec:ethics}

LLM-generated cultural heritage descriptions can shape how public audiences understand objects, places, and cultural histories. Our results show that fluent descriptions may introduce plausible but unsupported interpretations, especially for content fields such as \textit{Technique} and \textit{Subjects}. In public-facing settings, such errors risk presenting speculative claims as authoritative information.

This risk is particularly important for collections whose documentation is uneven across regions, periods, or communities. If models elaborate more confidently for some cultural-geographic contexts than others, generated descriptions may reinforce existing documentation imbalances rather than correct them. We therefore do not recommend using these systems as standalone curatorial tools. Instead, domain experts should review generated descriptions, and systems should show which claims are supported by catalogue metadata, which are inferred from images, and which remain uncertain.

Our work uses existing museum catalogue records and object images. We do not collect personal data or generate descriptions about living individuals. The main ethical concern is representational: ensuring that automated description generation does not overstate, distort, or homogenise cultural knowledge.

\bibliography{references}

\appendix
\input{latex/section-final/appendix}

\end{document}

%% file: latex/preamble/packages.tex
\usepackage{times}
\usepackage{latexsym}
\usepackage{amsmath}

\usepackage[T1]{fontenc}

\usepackage[utf8]{inputenc}

\usepackage{microtype}

\usepackage{inconsolata}
\usepackage[none]{hyphenat}

\usepackage{graphicx}

\usepackage{booktabs}      %
\usepackage{array}         %
\usepackage{multirow}      %
\usepackage{makecell}      %
\usepackage{caption}       %
\usepackage{threeparttable} %
\usepackage{xcolor}        %
\usepackage[table]{xcolor}
\usepackage{siunitx}
\usepackage{xspace}
\usepackage{subcaption}
\usepackage{amssymb}
\usepackage{enumitem}
\usepackage{hyperref}
\usepackage{url}
\usepackage{pdflscape}
\usepackage{booktabs}
\usepackage{ltablex}
\usepackage{array}
\keepXColumns

\usepackage{siunitx}
\usepackage[most]{tcolorbox}

\newtcolorbox[auto counter, number within=section]{promptbox}[2][]{
  colback=gray!4,
  colframe=gray!35,
  fonttitle=\small\bfseries,
  title={Prompt~\thetcbcounter: #2},
  label={prompt:#1},
  left=4pt,
  right=4pt,
  top=4pt,
  bottom=4pt,
  breakable
}

\usepackage{cleveref}

\newcommand{\bertscore}{BERTScore}

%% file: latex/preamble/authors.tex
\author{
  Clemencia Siro \and Laura Hollink \\
  Centrum Wiskunde \& Informatica (CWI) \\
  Amsterdam, The Netherlands \\
  \texttt{\{c.n.siro,l.hollink\}@cwi.nl}
}

%% file: latex/section-final/01-introduction.tex
\section{Introduction}
\label{sec:introduction}

Large language models (LLMs) are invreasingly used in cultural heritage settings to generate public-facing object descriptions from structured catalogue records~\cite{bu-etal-2025-investigation,trichopoulos2023large}. This is a metadata-conditioned data-to-text task: it requires fluent prose that stays faithful to a bounded, heterogeneous set of catalogue fields.
Data-to-text benchmarks such as E2E, WebNLG, RotoWire, and ToTTo have driven progress on content selection, fluency, and faithfulness~\citep{novikova-etal-2017-e2e,gardent-etal-2017-webnlg,wiseman-etal-2017-challenges, parikh-etal-2020-totto}, with evaluation focused on reference similarity, input coverage, and
hallucination~\citep{dusek-kasner-2020-evaluating,pagnoni-etal-2021-understanding}. However, this framing leaves unexplored how the composition of the input itself shapes generation.

Cultural heritage collections make this question especially important. Museum records are structured, but they are not uniform: catalogue fields differ in their epistemic status. Artefacts may encode provenance facts, physical descriptors, production dates, curatorial classifications, subject annotations, inscriptional content, and named associations within a single attribute set.
These fields differ in the extent to which a model can be expected to recover missing values from domain priors.
Multimodal models exacerbate this effect%
: an image supports description of form, colour, appearance, and visible markings~\cite{liu2024visual}, but cannot ground production place, excavation context, date, or associated historical figure. A model that conflates these grounding sources generates fluent prose that is, on the surface, indistinguishable from faithful description~\cite{maynez-etal-2020-faithfulness,ji-etal-2023-survey}. Field coverage is, in addition, systematically non-uniform across collections and cultural-geographic regions, reflecting heterogeneous acquisition histories and cataloguing conventions~\cite{srinivasan2017reimagining,vanhooland2014linked}.
As a result, metadata is not only an input representation for generation, but also a trace of prior documentation practice.

Prior work does not isolate these dimensions. Cultural-heritage evaluation often aggregates quality across examples or fields, and reports regional gaps observationally~\cite{ghaboura-etal-2025-time}, making it difficult to determine whether errors arise from metadata quantity, field identity, image availability, or domain-specific priors. Additionally, cultural heritage research has largely focused on improving access to collections, producing engaging descriptions, or linking heterogeneous museum records~\citep{hyvonen2012publishing,candela-etal-2023-colonial}, but less often examines how different metadata configurations change what models include, omit, or infer.

This paper asks: \emph{how do metadata quantity, field type, image availability, and place-specific documentation patterns affect what LLMs include, omit, and infer in museum description generation?} We address this question with a two-phase study on the TimeTravel benchmark~\citep{ghaboura-etal-2025-time}. Phase~1 provides a corpus-scale observational analysis over $10{,}250$ historical and cultural artefacts generated under each record's full available metadata. It shows that reference-based quality varies with metadata composition, but that naturalistic records confound field count, output length, and field composition: output length increases with field count ($\eta^2=0.33$), while longer generations receive lower BERTScore across civilisations ($r=-0.34$ to $-0.70$). Phase~2 introduces a \emph{controlled field-type probing framework} that independently varies field count, field type, and image availability within the same objects across ten experimental conditions. Two matched-count arm sequences extend from a common sparse baseline: the \textbf{context arm}~(\textsc{S\_ctx}) adds fields encoding the geographic, historical, and associative circumstances of the object (\textit{Find spot}, \textit{Place}, \textit{Assoc name}), and the \textbf{content arm}~(\textsc{S\_cont}) adds fields encoding properties directly perceivable from or readable on the object (\textit{Subjects}, \textit{Technique}, \textit{Inscription}). We evaluate each generation using reference-based surface similarity, NLI-estimated field coverage, and claim-level factuality via nugget extraction and verification.

Our results show that more metadata does not simply make generations more grounded. Increasing field count improves coverage of provided fields but can also increase unsupported inference over withheld fields. Field identity determines the dominant failure mode: contextual fields tend toward omission, while content fields are incorporated more readily but at higher hallucination risk. Images provide visual grounding in sparse settings but also encourage inference beyond the catalogue record. Place-level patterns further show that metadata effects depend on documentation practices and cultural-geographic priors, making field-level evaluation necessary for understanding LLM-generated museum descriptions.

We make three contributions. \textbf{(1)}~We introduce a \emph{controlled field-type probing framework} isolating the effects of metadata quantity, field type, and image access on LLM-generated museum descriptions. \textbf{(2)}~We show that more metadata does not improve faithfulness monotonically: increasing field count improves coverage of provided fields but increases unsupported inference over withheld fields, with contextual and content fields failing through distinct mechanisms. \textbf{(3)}~We show that place-level variation reflects field--region interactions and documentation practices, motivating field- and claim-level evaluation for metadata-conditioned generation.

%% file: latex/section-final/02-related-work.tex
\section{Related Work}
\label{sec:related-work}
\noindent
\textbf{Data-to-text generation}
Our work is situated in data-to-text generation, where
benchmarks such as E2E and WebNLG study generation from meaning representations and RDF triples, emphasising content selection, lexicalisation, aggregation, and surface realisation~\cite{novikova-etal-2017-e2e, gardent-etal-2017-webnlg}. Longer-form data-to-document tasks such as RotoWire show that generation from structured records also requires maintaining fidelity to the input database~\cite{wiseman-etal-2017-challenges}, while controlled table-to-text settings such as ToTTo restrict the input evidence through selected table cells to support more controlled evaluation of grounded generation~\cite{parikh-etal-2020-totto}. Museum description generation is a metadata-conditioned form of data-to-text generation, but with heterogeneous and incomplete records whose fields differ in availability and interpretive status.

\noindent
\textbf{Faithfulness in grounded generation}
A central challenge in grounded generation is that fluent outputs may include unsupported content. This problem has been widely studied in summarization, where models may generate hallucinated or factually inconsistent claims despite high surface quality~\cite{maynez-etal-2020-faithfulness, pagnoni-etal-2021-understanding}. Atomic claim-level verification has been proposed to improve factuality assessment beyond binary judgements~\cite{min-etal-2023-factscore}.
In data-to-text generation, factual grounding is similarly central: models may omit input facts, hallucinate unsupported content, or generate claims that cannot be traced to the source record~\cite{wiseman-etal-2017-challenges, wang-2019-revisiting}. We build on this line of work by analysing faithfulness at the metadata-field level.

\noindent
\textbf{Generation from cultural heritage data}
The use of LLMs on and for cultural heritage data is extensive. For example, ~\citet{bu-etal-2025-investigation} study how various open source LLMs answer questions about a museum exhibition; \citet{trichopoulos2023large} finetune an LLM with collection data to serve as a chatbot; \citet{arnold-tilton-2024-explainable} use LLMs to generate captions and explanations for image archives; \citet{reusens2025large} employ LLMs to generate keywords from written descriptions of museum objects. Generating natural language from museum collection databases has been studied in the past as a structured NLG
problem~\cite{isard-etal-2018-cycling}, demonstrating that metadata records can support adaptive visitor-facing
texts. In line with this, we generate descriptions based on structured records consisting of both images and metadata.

Metadata completeness is a central dimension of metadata quality \citep{bruce2004continuum,matoni2026qualitydescriptiveinformationcultural}. However, cultural heritage records are often incomplete. In Cultura Italia, for example, even mandatory fields are frequently missing \citep{LorenziniRospocherTonelli2021}. These gaps are not random. Metadata creation and aggregation reflect judgements about what information is valuable, which makes documentation practices subjective and culturally situated \citep{lewis-hyvonen2025}. As a result, completeness can vary systematically across subcollections, communities, or domains. This phenomenon is captured by Metadata Completeness Disparity \citep{Battaner-2026}.
To account for this, our work studies cultural heritage artefacts from various regions. By investigating how metadata completeness, image availability, and region interact and shape generation, we provide insights into whether models respect the boundary between verified catalogue information and model-supplied contextualisation.

%% file: latex/section-final/03-experiemtal-setup.tex
\section{Experimental Setup}
\label{sec:setup}

To understand how museum description generation is affected by the count and type of fields and whether visual support is available, we propose a
two-phase controlled field-type probing framework: Phase~1 measures generation under natural metadata variation, while Phase~2 applies controlled field probes that hold object identity fixed and vary metadata composition and image access.

\subsection{Task and Dataset}
\label{sec:dataset}

\noindent
\textbf{Task.} The task is structured metadata-conditioned museum description generation. A model $G_m$ receives a prompt $P(c, M_i, I_i)$ constructed from condition $c$, object metadata $M_i$, and optional image $I_i$, and generates a description: $\hat{y}_{i,m,c} = G_m\!\left(P(c,\, M_i,\, I_i)\right)$.
$\hat{y}_{i,m,c}$ is assessed against a reference description $y_i$. Fields provided in the prompt constitute \emph{primary fields} $F_\text{cond}$; fields present in the object record but withheld from the prompt are \emph{bonus fields} $F_\text{all} \setminus F_\text{cond}$. Primary-field analysis measures use of provided metadata; bonus-field analysis measures parametric inference over catalogue information absent from the prompt.

\noindent
\textbf{Dataset.}
We use the \textit{TimeTravel} benchmark~\cite{ghaboura-etal-2025-time}, comprising $10{,}250$ museum artefact records across ten geographic regions: %
Iraq, Egypt, Greece, Roman Empire, British Isles, China, Japan, Central America, India, and Iran. Each record contains structured catalogue metadata, an %
image, and a reference description. We analyse nine metadata fields: \textit{Materials}, \textit{Culture}, \textit{Production date}, \textit{Find spot}, \textit{Place}, \textit{Associated name}, \textit{Subjects}, \textit{Technique}, and \textit{Inscription}. %
The \textit{Place} field encodes which cultural-geographic region the artefact belongs to, and we use this to compare generations across regions.
The mean number of populated fields per object is $\bar{|F|}=6.8$ (range 3--9); documentation varies substantially across %
regions (Table~\ref{tab:metadata_availability1}).

\begin{table*}[t]
\centering
\small
\setlength{\tabcolsep}{3.5pt}

\begin{tabular}{l *{8}{S[table-format=3.1]}}
\toprule
\textbf{Place} & {\textbf{Assoc. name}} & {\textbf{Culture}} & {\textbf{Find spot}} & {\textbf{Inscription}} & {\textbf{Materials}} & {\textbf{Prod. date}} & {\textbf{Subjects}} & {\textbf{Technique}} \\
\midrule
\textbf{Overall} & 44.0 & 100.0 & 68.2 & 41.1 & 99.8 & 86.7 & 54.3 & 61.7 \\
\midrule
British Isles   & 1.0 & 100.0 & 98.0 & 1.0 & 100.0 & 93.0 & 6.0 & 66.0 \\
Central America & 1.0 & 100.0 & 100.0 & 0.0 & 100.0 & 22.0 & 3.0 & 25.0 \\
China           & 0.0 & 100.0 & 100.0 & 0.0 & 100.0 & 100.0 & 12.0 & 100.0 \\
Egypt           & 100.0 & 100.0 & 100.0 & 100.0 & 100.0 & 16.0 & 8.0 & 100.0 \\
Greece          & 61.0 & 100.0 & 86.0 & 16.0 & 100.0 & 100.0 & 100.0 & 89.0 \\
India           & 84.0 & 100.0 & 21.0 & 85.0 & 100.0 & 100.0 & 87.0 & 10.0 \\
Iran            & 18.0 & 100.0 & 21.0 & 51.0 & 98.0 & 99.0 & 59.0 & 88.0 \\
Iraq            & 28.0 & 100.0 & 94.0 & 30.0 & 100.0 & 100.0 & 87.0 & 41.0 \\
Japan           & 8.0 & 100.0 & 47.0 & 28.0 & 99.0 & 100.0 & 17.0 & 43.0 \\
Roman Empire    & 99.0 & 100.0 & 7.0 & 100.0 & 100.0 & 100.0 & 100.0 & 3.0 \\
\bottomrule
\end{tabular}
\caption{Percentage of metadata available per place and overall; overall availability is computed across all $10{,}250$ objects.}
\label{tab:metadata_availability1}
\end{table*}

\subsection{Two-Phase Experimental Design}
\label{sec:design}

\noindent
\textbf{Phase~1: natural metadata variation.}
Phase~1 evaluates generation under the metadata naturally available for each object. We prompt each model with the object's populated catalogue fields and image, generating descriptions for $10{,}250$ objects with four LLMs. This phase captures realistic collection-level variation, where field count, field type, image availability, and place co-vary and cannot be independently attributed to observed differences in generated descriptions, thus our motivating phase~2.

\noindent
\textbf{Phase~2: controlled field probes.}
We construct Phase~2 as a within-object design over $500$ objects balanced across values of the \textit{Place} %
field. Objects are sampled from place-relative high-completeness tiers, rather than a global completeness threshold, to avoid making metadata quantity a proxy for civilisation (Appendix~\ref{app:metadata-analysis}; Table~\ref{tab:field_count_distribution}). Each object appears under multiple prompt conditions (Table~\ref{tab:conditions}), so comparisons are paired within object. Object identity, and prompt template remain fixed, while field composition and image access vary.

\noindent
\textbf{Controlled generation conditions.}
\label{sec:conditions}
We define ten generation conditions (Table~\ref{tab:conditions}) that vary field count, field type, and image availability.
Condition~\textsc{S} is a three-field baseline comprising \textit{Materials}, \textit{Culture}, and \textit{Production date}, selected because each is populated for nearly all of the benchmark objects. Six of the ten conditions form a 2 (field type) $\times$ 3 (field count) design over this baseline: each arm extends \textsc{S} by one field per level, so that level $k$ is a superset of level $k-1$.
The two arms partition the added fields along the content--context distinction~\citep{gilliland2016setting}: content comprises properties intrinsic to the object, including its subject matter, means of production, and any text it bears; context comprises properties extrinsic to the object, recording the circumstances of its creation, provenance, and custodial history.

The \textbf{context arm} (\textsc{S\_ctx}4--6) adds \textit{Find spot}, then \textit{Place}, then \textit{Assoc name}, which link the object to external documentary and associative records.
The \textbf{content arm} (\textsc{S\_cont}4--6) adds \textit{Subjects}, then \textit{Technique}, then \textit{Inscription}, which encode properties observable in the object or inscribed on it, although assigning catalogue values to them requires expert interpretation.
Field order within each arm is fixed across all objects and models, so level $k$ denotes an identical field set for every generation.
The arms are matched on field count at levels 4, 5, and 6, and object identity, prompt template, and image availability are held fixed across matched pairs, so differences at a given level are attributable to field type rather than field count.
The remaining four conditions: \textsc{R} and \textsc{T} bound the field range, and \textsc{S\_noi} and \textsc{R\_noi} withhold the image.

\begin{table}[t]
\centering\small
\setlength{\tabcolsep}{3pt}
\begin{tabular}{llcc}
\toprule
\textbf{Cond.} & \textbf{Fields provided} & $|F|$ & \textbf{Img.}\\
\midrule
\textsc{R}       & All available                               & 6--9 & \checkmark \\
\textsc{T}       & All available                               & 6--9 & \texttimes \\
\textsc{S}       & Materials, Culture, Prod.\ date             & 3    & \checkmark \\
\textsc{S\_noi}  & Materials, Culture, Prod.\ date             & 3    & \texttimes \\
\midrule
\textsc{S\_ctx4} & \textsc{S} + Find spot                      & 4    & \checkmark \\
\textsc{S\_ctx5} & \textsc{S} + Find spot, Place               & 5    & \checkmark \\
\textsc{S\_ctx6} & \textsc{S} + Find spot, Place, Assoc.\ name & 6    & \checkmark \\
\midrule
\textsc{S\_cont4} & \textsc{S} + Subjects                       & 4    & \checkmark \\
\textsc{S\_cont5} & \textsc{S} + Subjects, Technique            & 5    & \checkmark \\
\textsc{S\_cont6} & \textsc{S} + Subjects, Technique, Inscr.    & 6    & \checkmark \\
\bottomrule
\end{tabular}
\caption{Phase~2 controlled prompt conditions. Matched-count arm conditions compare field type while holding object identity, prompt template, and image access fixed.}
\label{tab:conditions}
\end{table}

\subsection{Evaluation}
\label{sec:evaluation}

We evaluate the generations on three dimensions:

\noindent
\textbf{Reference similarity.}
We compute BLEU-1~\citep{papineni-etal-2002-bleu}, ROUGE-L~\cite{lin-2004-rouge}, METEOR~\cite{banerjee-lavie-2005-meteor}, and BERTScore (\texttt{roberta-large})~\cite{zhang2020bertscore} against $y_i$.

\noindent
\textbf{Field-value support.}
We measure whether $\hat{y}_{i,m,c}$ supports catalogue field values using DeBERTa-large-MNLI~\cite{he2021deberta}. For each $f \in F^{\mathrm{eff}}_{i,c}$ with value $v_f$, we construct a value-specific hypothesis $h_f$  and classify $\hat{y}$ via argmax over \texttt{entailment}, \texttt{neutral}, and \texttt{contradiction}~\footnote{We list the hypothesis for each field category in Appendix~\ref{app:evaluation} (Table~\ref{tab:nli_templates})}. Primary coverage is the entailment rate over the provided fields.
For bonus fields, entailment indicates model recovery of a withheld catalogue value; contradiction indicates incorrect inference.

\noindent
\textbf{Claim-level factuality.}
NLI provides one entailment verdict per field and cannot identify which claims in the generation are responsible for that verdict. We therefore apply a two-stage nugget pipeline to verify factual accuracy at the claim level. Stage~1 decomposes $g$ into atomic claims, attributing each to a field $f \in F_\text{all}$ or classifying it as \emph{extra content} when no schema field accounts for the claim.

Stage~2 verifies each attributed claim against $F_\text{all}$, assigning \texttt{supported}, \texttt{hallucinated}, or \texttt{not\_verifiable} with error-type classification~(\Cref{tab:error_taxonomy}). Both labels are defined relative to the catalogue record rather than to external fact: a claim is \texttt{hallucinated} when the field to which it is attributed has a recorded value and the claim is not supported by that value, and \texttt{not\_verifiable} when the field has no recorded value against which to assess it. A claim may therefore be historically accurate or visually correct and still be labelled \texttt{hallucinated}. Pipeline validation is reported in Appendix~\ref{app:validation-checks}.

The primary hallucination rate $\mathrm{HallP}(i,c)$ is the proportion of claims in $\mathcal{N}^P_{i,c}$ assigned \texttt{hallucinated} by Stage~2. Claim verdicts are aggregated to the field level: a primary field $f \in F_\text{cond}$ is \texttt{hallucinated} if any attributed claim is hallucinated, \texttt{covered} if at least one is supported and none are hallucinated, and \texttt{silent} if no claims are attributed to it. Bonus fields $f \in F_\text{all} \setminus F_\text{cond}$ follow the same logic with labels \texttt{correct}, \texttt{wrong}, and \texttt{silent}. The extra-content rate is the proportion of claims unattributable to any field in $F_\text{all}$.

Implementation details, including model versions, prompt templates, and significance tests, are provided in Appendix~\ref{app:evaluation}.

\smallskip
\noindent\textbf{Human calibration.}
We assess the reliability of the claim-level pipeline against human annotation.
Two annotators independently labeled a stratified random sample of 350 claims, drawn proportionally across conditions, following an annotation protocol derived from the Stage~2 prompt.
Inter-annotator reliability is substantial (Cohen's $\kappa = 0.78$), with category-specific agreement of 96\% for \texttt{supported} and 87\% for \texttt{hallucinated}, showing that human raters apply the verdict criterion consistently without reference to the pipeline.
Human--pipeline reliability is $\kappa_w = 0.67$ with category-specific agreement of 89\% for \texttt{supported} and 73\% for \texttt{hallucinated}. This shows that our nugget pipeline is able to correctly attribute most nuggets to their specific fields.

%% file: latex/section-final/04-results.tex
\section{Results}
\label{sec:results}
We report results across the two phases.

\subsection{Phase~1: Surface metrics invert documentation completeness}
\label{sec:phase1}

\begin{figure}[t]
    \centering
    \includegraphics[width=1\columnwidth]{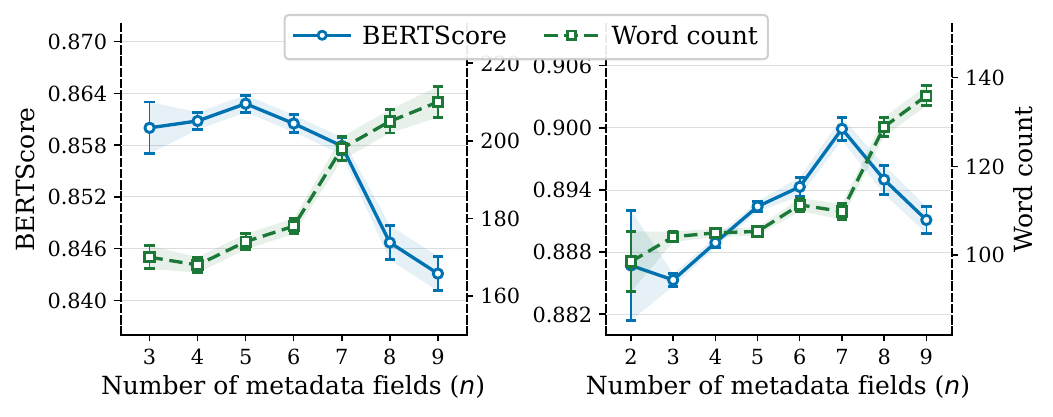}
    \caption{Mean BERTScore and generated word count as a function of the number of populated metadata fields~$n$, across Phase~1 (left) and Phase~2 (right).}
    \label{fig:bertscore_wordcount_phases}
\end{figure}

Phase~1 generates descriptions under each artefact's full available metadata. Figure~\ref{fig:bertscore_wordcount_phases}(a) shows that mean description length increases monotonically with field count, from 160 to 213 words across the range of 3--9 fields, while mean BERTScore-P decreases from 0.863 to 0.839 over the same range. Across regions, BERTScore is consistently and negatively associated with description length, with correlations ranging from $r=-0.34$ to $r=-0.70$s ($\eta^2 = 0.33$ for the length--field-count relationship~\Cref{tab:regional_effects}): indicating that more fields correspond to longer generations, which receive lower BERTScore against the per-object reference description.

Surface-metric rankings almost perfectly invert documentation completeness (Kendall's~$W = 0.990$), with well-documented regions scoring lower despite richer records (Table~\ref{tab:bertscore_by_place}), replicating the regional difficulty pattern reported by~\citet{ghaboura-etal-2025-time}.\footnote{Per-model results across all surface metrics are reported in Appendix~\ref{app:phase1-results}.} As field count and region co-vary in naturalistic data, their individual contributions to quality variation cannot be disentangled from observational results. Below, we report our results from phase~2, which disentangles these dimensions.

\subsection{Metadata composition creates grounding trade-offs}
\label{sec:rq1}
\noindent
\textbf{Field count scales bonus inference without improving primary grounding.}
Figure~\ref{fig:bertscore_wordcount_phases}(b) shows that in Phase~2, BERTScore increases from 0.885 at $n = 3$ to 0.901 at $n = 7$, before decreasing to 0.890 at $n = 9$, while word count increases monotonically from 100 to 135 words. Unlike Phase~1, the inverse relationship between description length and BERTScore does not hold uniformly once object identity is held fixed. Table~\ref{tab:corr_place_nfields_length} confirms this at the regional level: within-place field-count--BERTScore correlations shift from negative across all regions in Phase~1 to positive across all ten %
regions in Phase~2 ($\rho = 0.200$ to $0.421$).

\begin{figure}
    \centering
    \includegraphics[width=\columnwidth]{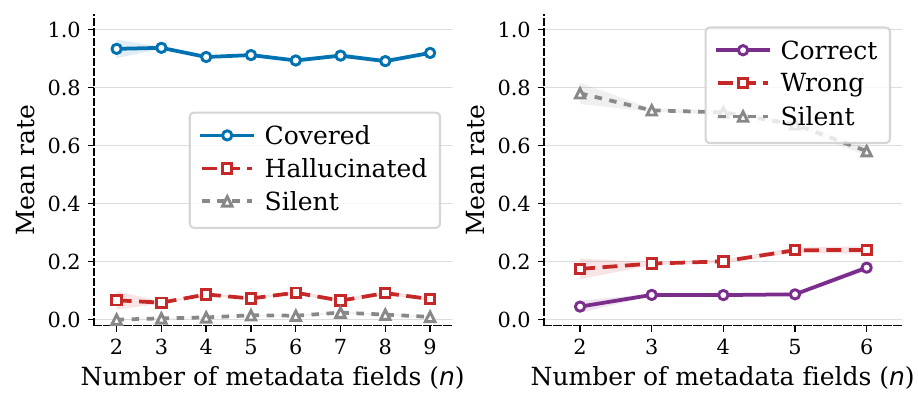}
    \caption{Mean primary fields'~(coverage, hallucination, silence; left) and bonus fields'~(correct, wrong, silent; right) as a function of observed metadata field count~$n$.
}

    \label{fig:field_coverage_by_n}
\end{figure}

Field-count effects in Phase~2 show that increasing metadata quantity does not uniformly improve grounding, as shown in \Cref{fig:field_coverage_by_n}. For primary fields, coverage remains stable across all field counts, ranging from $0.90$ to $0.93$, while hallucination rates vary between $0.07$ and $0.10$. Silence rates remain near zero throughout, indicating that models verbalise provided fields regardless of how many are provided.

Bonus field rates show greater variation with field count. Bonus silence decreases from $0.79$ at $n = 2$ to $0.65$ at $n = 6$, while both correct and wrong bonus rates increase over the same range. Bonus wrong exceeds bonus correct at every observed field count, with wrong rates stable at $0.18$--$0.24$ against correct rates rising from $0.05$ to $0.19$. Higher field counts are therefore associated with both greater parametric recovery of withheld fields and greater metadata-boundary overreach.

\noindent
\textbf{Field type changes the dominant failure mode.}
The claim-level pipeline reveals systematic differences between matched-count conditions. \textsc{S\_ctx} conditions are more conservative and more faithful on primary fields: \textsc{S\_ctx4--6} achieve primary coverage of $0.937$--$0.949$ with hallucination rates of $0.040$--$0.044$ (Table~\ref{tab:halluc_by_condition}). \textsc{S\_cont} conditions produce lower primary coverage ($0.867$--$0.893$) and higher hallucination ($0.102$--$0.129$). The largest contrast occurs in \textsc{S\_cont5}, which adds \textit{Subjects} and \textit{Technique}: it has the highest primary hallucination rate ($0.129$) while maintaining near-zero silence ($0.004$).
\textsc{S\_cont} fields therefore show high primary coverage rates while also having high primary hallucination rates.
Paired permutation tests over matched field counts (10{,}000 resamples, BH--FDR corrected) confirm that this split holds at every step of the arm: \textsc{S\_ctx} exceeds \textsc{S\_cont} in primary coverage by $5.6$--$7.4$~pp and in bonus wrong by $15.5$--$34.5$~pp, and falls below it in primary hallucination by $6.2$--$8.5$~pp (all $p < 0.001$).

\begin{table}[!t]
\centering
\small
\setlength{\tabcolsep}{4pt}
\begin{tabular}{lccc|ccc}
\toprule
&
\multicolumn{3}{c|}{\textbf{Primary fields}} &
\multicolumn{3}{c}{\textbf{Bonus fields}} \\
\cmidrule(lr){2-4}\cmidrule(l){5-7}
\textbf{Cond.} & Cov & Hall & Sil & Cor & Wro & Sil\textsubscript{b} \\
\midrule
\textsc{S}
  & 0.933 & 0.058 & 0.009
  & 0.092 & 0.223 & 0.685 \\
\textsc{S\_noi}
  & 0.885 & 0.091 & 0.024
  & 0.055 & 0.146 & 0.799 \\
\textsc{S\_ctx4}
  & \textbf{0.949} & \textbf{0.040} & 0.011
  & 0.116 & 0.308 & 0.577 \\
\textsc{S\_ctx5}
  & 0.941 & 0.044 & 0.016
  & 0.128 & 0.361 & 0.511 \\
\textsc{S\_ctx6}
  & 0.937 & 0.042 & 0.021
  & \textbf{0.218} & 0.421 & 0.360 \\
\textsc{S\_cont4}
  & 0.893 & 0.102 & 0.005
  & 0.062 & 0.153 & 0.786 \\
\textsc{S\_cont5}
  & 0.867 & 0.129 & \textbf{0.004}
  & 0.032 & \textbf{0.031} & \textbf{0.937} \\
\textsc{S\_cont6}
  & 0.876 & 0.113 & 0.011
  & 0.060 & 0.076 & 0.864 \\
\textsc{R}
  & 0.885 & 0.092 & 0.023
  & --- & --- & --- \\
\textsc{T}
  & 0.879 & 0.082 & 0.039
  & --- & --- & --- \\
\bottomrule
\end{tabular}
\caption{Primary fields rates (Cov, Hall, Sil) and bonus completion rates (Cor, Wro, Sil\textsubscript{b}) by condition, averaged across all models; Bold indicates best value per column: highest for Cov, Cor, Sil\textsubscript{b}; lowest for Hall, Sil, Wro. \textsc{R} and \textsc{T} contain no withheld fields by design.}
\label{tab:halluc_by_condition}
\end{table}

Table~\ref{tab:field_outcomes} shows that, \textit{Subjects} and \textit{Technique} have the highest primary hallucination rates ($24.2\%$ and $21.3\%$), far above fields such as \textit{Culture} ($1.3\%$) and \textit{Production date} ($1.7\%$). \textsc{S\_ctx} fields have lower hallucination but higher omission: \textit{Find spot} and \textit{Place} show hallucination rates of $3.9\%$ and $5.6\%$, with silence rates of $5.0\%$ and $8.1\%$. \textsc{S\_cont} fields therefore fail primarily through unsupported elaboration, whereas \textsc{S\_ctx} fields fail primarily through omission. \textit{Inscription} is the only field for which bonus correct exceeds bonus wrong ($16.7\%$ vs.\ $10.6\%$), alongside a moderate primary hallucination rate ($9.1\%$), indicating that \textit{Inscription} content is partially recoverable when withheld.

Withheld-field follows the same field-type split. Wrong bonus mentions exceed correct ones overall, indicating that unprompted field recovery is more often speculative than reliable. The highest bonus wrong rates occur for \textit{Technique} ($51.7\%$) and \textit{Subjects} ($41.8\%$). In the \textsc{S\_ctx} arm, where \textsc{S\_cont} fields remain withheld, bonus wrong increases from $0.308$ in \textsc{S\_ctx4} to $0.421$ in \textsc{S\_ctx6}, while bonus silence decreases from $0.577$ to $0.360$ (Table~\ref{tab:halluc_by_condition}). Additional context from \textsc{S\_ctx} increases completion of the withheld record, but much of this completion is unsupported. In contrast, withheld \textsc{S\_ctx} and association fields are usually omitted: \textit{Find spot} has $99.2\%$ bonus silence and $0.4\%$ bonus wrong, while \textit{Place} and \textit{Assoc. name} show bonus silence rates of $86.8\%$ and $88.0\%$. Field type, therefore, determines whether missing information is omitted or speculatively completed, making field count alone insufficient for explaining claim-level faithfulness.
The gap widens along the \textsc{S\_ctx} arm, from $15.5$~pp at count~4 to $34.5$~pp at count~6, while bonus silence diverges in the opposite direction, from $0.577$ against $0.786$ to $0.360$ against $0.864$ (all $p < 0.001$).

\begin{table}[!t]
\centering\small

\begin{tabular}{lcccccc}
\toprule
& \multicolumn{3}{c}{\textbf{Primary}}
& \multicolumn{3}{c}{\textbf{Bonus}} \\
\cmidrule(lr){2-4}\cmidrule(lr){5-7}
\textbf{Field} & Cov & Hall & Sil & Cor & Wro & Sil\textsubscript{b} \\
\midrule
\textit{Subjects}        & 74.9          & 24.2          & 0.9          & 14.6          & 41.8          & 43.6          \\
\textit{Technique}       & 78.3          & 21.3          & \textbf{0.3} & 9.8           & 51.7          & 38.5          \\
\textit{Assoc name}      & 87.1          & 11.9          & 1.0          & 5.5           & 6.5           & 88.0          \\
\textit{Materials}       & 88.7          & 10.9          & 0.4          & ---           & ---           & ---           \\
\textit{Inscription}     & 87.5          & 9.1           & 3.4          & \textbf{16.7} & 10.6          & 72.6          \\
\textit{Place}           & 86.3          & 5.6           & 8.1          & 6.9           & 6.3           & 86.8          \\
\textit{Find spot}       & 91.2          & 3.9           & 5.0          & 0.4           & \textbf{0.4}  & \textbf{99.2} \\
\textit{Prod. date} & 96.2          & 1.7           & 2.1          & ---           & ---           & ---           \\
\textit{Culture}         & \textbf{98.2} & \textbf{1.3}  & 0.5          & ---           & ---           & ---           \\
\bottomrule
\end{tabular}
\caption{Primary fields rates (Cov, Hall, Sil) and bonus completion rates (Cor, Wro, Sil\textsubscript{b}) by metadata field, averaged across all conditions and models.}
\label{tab:field_outcomes}
\end{table}

\subsection{Grounding reliability varies by region}
\label{sec:rq3}

\noindent
\textbf{Regions differ in dominant failure mode, not in overall reliability.}
BERTScore varies significantly across regions in all conditions (Kruskal--Wallis $H > 100$, $p \ll 0.001$; Table~\ref{tab:bertscore_place_model}), but over a narrow range of means, from $0.877$ (Greece) to $0.896$ (Central America).
A negative length--BERTScore correlation holds within every region ($r = -0.38$ to $-0.71$, all $p \ll 0.001$; Figure~\ref{fig:bertscore_wc_places}).
Claim-level rates separate the regions more sharply than BERTScore does (Table~\ref{tab:nugget_by_place}): the British Isles records the highest primary coverage ($0.936$) and lowest hallucination ($0.054$), Central America the highest hallucination ($0.112$), and China the lowest coverage ($0.859$) with the highest silence ($0.047$).
Bonus wrong exceeds bonus correct in every region; India records the lowest bonus wrong rate ($0.133$) and the Roman Empire the highest bonus correct ($0.170$).

These aggregate differences arise from field--region interactions (Figure~\ref{fig:field_place_3x2_full}).
\textit{Subjects} and \textit{Technique} record the highest primary hallucination rates in every region, but the magnitude varies by a factor of $2$--$4$: among regions where the field is populated, \textit{Technique} ranges from $0$ (Roman Empire) to $0.46$ (Central America), and \textit{Subjects} spans a narrower $0.14$--$0.35$.
China and Iran record higher primary silence on \textsc{S\_ctx} fields than other regions (\textit{Find spot} $0.26$ and \textit{Place} $0.24$ for China; \textit{Find spot} $0.19$ for Iran).
For Greece, models generate bonus \textit{Technique} content in $71\%$ of cases, of which $0.69$ are wrong against $0.04$ correct.

Part of the apparent regional variation reflects catalogue composition rather than generation behaviour.
The near-zero \textit{Technique} hallucination rate for the Roman Empire follows from the field being absent from its records, so no primary claims are exposed to verification against it.
Regional rates are therefore conditional on which fields each collection populates, and comparisons across regions hold only at matched field type and field count.

Generation failure is not uniform across regions.
The British Isles exhibits high faithfulness to the provided metadata; Central America produces unsupported elaboration on \textit{Subjects} and \textit{Technique}; China omits provided field content; Greece incorrectly infers withheld \textit{Technique}.
These modes carry different consequences for museum description: hallucination introduces false content, silence omits documented content, and unsupported bonus inference fabricates attributes absent from the record.
Deployment in cultural heritage settings therefore requires region-specific quality assurance rather than a single global accuracy threshold.

\begin{table}[t]
\centering
\small
\setlength{\tabcolsep}{4pt}
\begin{tabular}{lccc|ccc}
\toprule
&
\multicolumn{3}{c|}{\textbf{Primary fields}} &
\multicolumn{3}{c}{\textbf{Bonus fields}} \\
\cmidrule(lr){2-4}\cmidrule(l){5-7}
\textbf{region} &
Cov & Hall & Sil & Cor & Wro & Sil\textsubscript{b} \\
\midrule
\textit{Central A.} & 0.877 & 0.112 & 0.010 & 0.050 & 0.304 & 0.646 \\
\textit{China}           & 0.859 & 0.094 & \textbf{0.047} & 0.062 & 0.268 & 0.669 \\
\textit{British I.}   & \textbf{0.936} & \textbf{0.054} & 0.010 & 0.059 & 0.232 & 0.710 \\
\textit{Japan}           & 0.918 & 0.072 & 0.010 & 0.114 & 0.185 & 0.701 \\
\textit{India}           & 0.901 & 0.082 & 0.017 & \textbf{0.137} & \textbf{0.133} & 0.730 \\
\textit{Iraq}            & 0.897 & 0.093 & 0.010 & 0.113 & 0.187 & 0.700 \\
\textit{Iran}            & 0.913 & 0.068 & 0.020 & 0.102 & 0.199 & 0.699 \\
\textit{Egypt}           & 0.922 & 0.063 & 0.015 & 0.082 & 0.145 & \textbf{0.774} \\
\textit{Roman E.}    & 0.912 & 0.071 & 0.017 & 0.170 & 0.236 & 0.593 \\
\textit{Greece}          & 0.898 & 0.086 & 0.016 & 0.051 & 0.234 & 0.715 \\
\bottomrule
\end{tabular}
\caption{Primary and bonus fields results from the claim-level factuality.
}
\label{tab:nugget_by_place}
\end{table}

\subsection{Image helps sparse grounding but increases extrapolation}
\label{sec:rq2}

\noindent
\textbf{Visual input affects grounding only under sparse metadata.}
Image ablation under sparse metadata (\textsc{S\_noi} versus \textsc{S}) reduces BERTScore across all models, lowers primary coverage from $93.3\%$ to $88.5\%$, and raises primary hallucination from $5.8\%$ to $9.1\%$ (\Cref{tab:halluc_by_condition}).
Paired permutation tests find each difference significant ($4.8$~pp coverage, $3.3$~pp hallucination, $7.7$~pp bonus wrong; all $p < 0.001$).
The same ablation under full metadata yields no significant difference: \textsc{R} and \textsc{T} differ by $0.6$~pp in primary coverage and $1.0$~pp in primary hallucination, neither significant after correction.
The contribution of visual input to primary grounding is therefore conditional on metadata density, and is not detectable once the prompt specifies all available fields.

The effect is concentrated in withheld rather than provided fields.
Ablation raises bonus silence from $68.5\%$ to $79.9\%$ and lowers bonus wrong from $22.3\%$ to $14.6\%$, with the wrong-to-correct ratio increasing from $2.4$:$1$ to $2.7$:$1$.
Image-inclusive conditions thus produce more attempts at withheld fields and a higher absolute count of incorrect completions, but a lower proportion of incorrect attempts.
Visual input extends generation beyond the provided fields without a corresponding gain in accuracy, reproducing the metadata-boundary overreach observed along the \textsc{S\_ctx} arm (\Cref{sec:rq1}) from a visual rather than a textual source.

Model-level effects vary under full metadata.
\textsc{LLaMA-Vision} records the largest \textsc{R}-to-\textsc{T} primary coverage difference of any model ($0.882$ to $0.796$; $\Delta = 0.086$), against an aggregate difference of $0.6$~pp.
Per-model and per-region results across field count, type, place, and image ablation are reported in Appendix~\ref{app:results}.

%% file: latex/section-final/05-discussion.tex
\section{Discussion}
\label{sec:discussion}

Our results show that metadata-conditioned generation is not only a question of input coverage, but also of field--claim alignment. Adding metadata changes what models mention, but also how they redistribute information across catalogue fields, visual evidence, and domain priors. This challenges the prevailing data-to-text assumption that richer inputs consistently improve generation quality.

First, metadata quantity is not a reliable indicator of grounding. In Phase~2, increasing field count improves coverage of provided fields but also increases hallucination over withheld fields. Better grounding of visible metadata can therefore coincide with weaker respect for the metadata boundary. Additional fields change the distribution of claims in the output: models mention more catalogue-supported information, but they also produce longer descriptions with more opportunities for unsupported elaboration. This distinction is visible in the metric results. Field-count correlations with \bertscore{} vary across regions, whereas output length is consistently and negatively associated with \bertscore{}. Surface metrics, therefore, risk penalizing grounded elaboration while rewarding shorter descriptions that are closer to the reference but less complete.

Second, field type shapes the primary error distribution. Context fields (\textit{Find spot}, \textit{Place},
\textit{Assoc name}), are reliably grounded when present and are rarely inferred when absent. Whereas content fields such as \textit{Technique} and \textit{Subjects} encourage domain-specific elaboration and visual inference, resulting in higher hallucination rates even when provided. \textit{Materials} occupies an intermediate position: it is usually grounded when present, but can trigger unsupported object-type or quality inferences. The context arm (\textsc{S\_ctx}), therefore, improves primary fidelity, while the content arm (\textsc{S\_cont}) increases primary coverage at higher hallucination cost. These patterns show that many errors are not arbitrary fabrications. Instead, models transfer values across neighbouring fields, narrow broad catalogue labels using domain priors, or convert visual observations into field-specific claims. The error often lies not in whether a claim is plausible, but in whether the correct field supports it.

This field--claim mismatch is the central failure mode our analysis reveals. A generated statement may be visually plausible, historically reasonable, or supported by another metadata field, while still being unsupported by the target field being evaluated. For example, a model may use a \textit{Find spot} value as a \textit{Place} of origin, convert a broad technique label into a specific art-historical sub-type, or add subject details visible in the image but absent from the catalogue subject vocabulary. Such cases are difficult to capture with evaluation methods that only check whether a claim is plausible or whether it appears somewhere in the input. Metadata-conditioned generation requires verifying the relation between claims and the specific fields that support them. Omission and hallucination also carry different consequences: omission reduces completeness, whereas unsupported elaboration presents unverified claims as catalogue-grounded facts.

Images further expose this issue, and their effect depends on metadata density. In sparse conditions, images provide grounding evidence that reduces hallucination; under richer metadata, they can amplify content-field elaboration beyond the catalogue record~\citep{rohrbach-etal-2018-object}. Our evaluation treats catalogue metadata as the reference schema, not as an exhaustive description of the object. Image-grounded claims may therefore be visually valid while remaining unsupported by the metadata field under evaluation. This distinction is important for cultural heritage generation: a description can be visually plausible while going beyond what the catalogue schema supports.

These findings have implications for evaluating metadata-conditioned generation. Reference similarity metrics, field-level NLI, and claim-level verification capture different aspects of quality and can produce conflicting condition rankings. NLI-based coverage measures
semantic consistency rather than value accuracy: a description can receive an entailment verdict without correctly reproducing the
catalogue value~\citep{falke-etal-2019-ranking}, and no single evaluation layer separates omission from hallucination. Evaluation
should therefore be field-aware: not only whether a claim is supported, but which field supports it, whether that field was
provided, and whether the claim stays within its intended scope. Treating all metadata fields as equivalent hides important differences in coverage rate, hallucination risk, and omission. More broadly, metadata-conditioned generation should be treated as a schema-alignment problem: models must learn not only what information is available, but what each field permits them to assert. In cultural heritage settings, unsupported elaboration can present speculative interpretation as authoritative descriptions.

%% file: latex/section-final/06-conclusion.tex
\section{Conclusion}
\label{sec:conclusion}

We presented a controlled study of metadata-conditioned description generation for museum objects with multimodal LLMs, varying the number of metadata fields, their type, and image availability over a fixed object set.
Increasing the number of provided fields does not monotonically improve faithfulness: entailment-based coverage of the provided fields rises, and so does the rate of unsupported claims about fields withheld from the prompt. Field type affects both quantities at a fixed number of fields.
Context fields are either expressed or omitted and are rarely misstated, whereas content fields such as \textit{Technique} and \textit{Subjects} are expressed whenever provided but yield the highest rate of unsupported claims and encourage visual inference, domain-specific elaboration, and schema-boundary errors. Reference-based similarity and entailment-based coverage do not distinguish these behaviours: they order the matched conditions differently, and a difference in error type is read as a difference in quality.
Faithfulness requires attributing each generated claim to the metadata field it addresses and verifying it against that field's value, which separates support, contradiction, and omission. For cultural heritage applications, this matters because fluent descriptions can present plausible but unsupported interpretation as authoritative object information. Systems deployed in this setting should therefore expose the provenance of each claim, so that catalogue-supported content can be reviewed separately from content generated from parametric knowledge.

%% file: latex/section-final/appendix.tex
\section{Experimental Setup Details}
\label{app:methodology-details}

This appendix provides additional detail on metadata field availability and the place-relative sampling procedure used to construct the Phase~2 controlled sample. We also discuss how we handled field collapse across conditions and list the prompts used in generation and evaluation.

\subsection{Metadata Availability and Place-Relative Sampling}
\label{app:metadata-analysis}

Table~\ref{tab:metadata_availability1} reports field availability by region, defined as the proportion of objects within each region for which a given field is populated. \textit{Culture}, \textit{Place}, and \textit{Materials} are near-universally populated across all regions. Some fields vary substantially: \textit{Subjects} is fully populated for Greece and the Roman Empire but below $10\%$ for British Isles and Central America; \textit{Find spot} is fully populated for Central America and China but below $25\%$ for India and the Roman Empire; \textit{Inscription} ranges from $0\%$ in Central America and China to $100\%$ in Egypt. These disparities reflect differences in institutional documentation practice across collections and motivate the place-relative sampling for phase~2.

In phase~2, we apply a place-relative criterion: for each region, we sample $50$ objects from the highest available field-count tiers within that category, defined as the one or two tiers with the largest object counts at the upper end of each region's completeness distribution.
We refrain from using a global completeness threshold as this would disproportionately exclude regions with lower field counts, biasing the controlled sample toward better-documented regions. This ensures all ten regions are represented in phase~2 and that the ceiling conditions (\textsc{R} and \textsc{T}) use the richest metadata available within each region rather than a globally imposed minimum.

Table~\ref{tab:field_count_distribution} shows the distribution of objects by region and field count. Bold cells indicate the tiers from which Phase~2 objects are drawn. The minimum selected tier ranges from six fields for Central America, where nine-field objects are absent or negligible, to seven or more for the remaining regions.

\begin{table*}[t]
\centering
\setlength{\tabcolsep}{6pt}
\begin{tabular}{lrrrrrrr}
\toprule
\textbf{Place} & \textbf{3} & \textbf{4} & \textbf{5} & \textbf{6} & \textbf{7} & \textbf{8} & \textbf{9} \\
\midrule
British Isles   & 17 & 45 & 276 & 616 & \textbf{54} & \textbf{5} & -- \\
Iran             & 15 & 18 & 155 & 352 & 332 & \textbf{125} & \textbf{2} \\
Japan            & 5 & 239 & 225 & 200 & \textbf{180} & \textbf{20} & -- \\
Central America  & -- & 304 & 132 & \textbf{48} & \textbf{3} & \textbf{1} & -- \\
India            & -- & 7 & 77 & 50 & 621 & \textbf{112} & 1 \\
Iraq             & -- & -- & 34 & 400 & 355 & 159 & \textbf{52} \\
Greece           & -- & -- & 14 & 100 & 861 & 709 & \textbf{185} \\
China            & -- & -- & -- & 918 & \textbf{117} & \textbf{3} & \textbf{1} \\
Roman Empire     & -- & -- & -- & 12 & 943 & \textbf{88} & \textbf{6} \\
Egypt            & -- & -- & -- & -- & 861 & 140 & \textbf{55} \\
\bottomrule
\end{tabular}
\caption{Distribution of objects by \textit{Place} and number of available metadata fields. Bold cells indicate place-relative high-completeness tiers used for Phase~2 sampling.}
\label{tab:field_count_distribution}
\end{table*}

\subsection{Collapsed Conditions}
\label{app:collapse1}

A condition is \emph{collapsed} for object $i$ when $|F^{\mathrm{eff}}_{i,c}| < |F^{\mathrm{tar}}_c|$, occurring when one or more target fields are unpopulated in the object catalogue record. Collapse is most frequent in \textsc{S\_cont6} (\textit{Inscription}: $41\%$ collection-wide availability) and \textsc{S\_str6} (\textit{Associated name}: $44\%$). All field-level analyses use $F^{\mathrm{eff}}_{i,c}$ throughout; collapsed instances are retained in aggregate metrics.

To verify robustness, we recompute $\mathrm{CovP}$ and $\mathrm{HallP}$ on the non-collapsed subset for the three conditions with the highest collapse rates. Mean absolute deviation from full-sample estimates is below $0.01$ in all cases, confirming that collapsed instances do not drive reported patterns.

\subsection{Models and Implementation Details}
\label{app:evaluation}

\paragraph{Generation models.}
Table~\ref{tab:model-versions} lists the exact model versions used in the study. All generation models are run with temperature~0 and a maximum output length of 512~tokens. No output reached this limit. The fixed generation prompt is shown in Prompt~\ref{prompt:generation}. In Phase~2, LLaMA-3.2-Vision is run only on the image ablation conditions~(\textsc{R}, \textsc{S}, \textsc{S\_noi}, and \textsc{T}). The other three models are run on all ten conditions, yielding 17{,}000 attempted generations. In Phase~1, all four models generate descriptions for all 10{,}250 objects under full metadata, giving $n=41{,}000$ generations. GPT-4o is accessed via OpenAI, while the remaining generation models are accessed via DeepInfra. Fewer than five failed or empty Phase~2 responses are excluded.

\paragraph{Image input.}
For image-enabled conditions, the artefact photograph is appended to the user prompt as a base64 JPEG vision block. We apply no client-side resizing or preprocessing. Provider APIs handle any internal normalisation. No image is included for \textsc{S\_noi} or \textsc{T}.

\paragraph{Surface metrics.}
BLEU-1 is computed with \texttt{nltk} using smoothing method~1. ROUGE-L is computed with \texttt{rouge-score} and stemming. METEOR is computed with \texttt{nltk.translate.meteor\_score}. BERTScore is computed with \texttt{roberta-large}, no IDF weighting, and no baseline rescaling.
BERTScore precision ($P_{\text{BERT}}$) is the primary surface metric.

\paragraph{NLI field coverage.}
Field coverage is computed with \texttt{microsoft/deberta-large-mnli} \citep{he2021deberta}. Each field-value pair is converted into a template hypothesis (Table~\ref{tab:nli_templates}) and classified against the generated description by argmax over entailment, neutral, and contradiction. Entailment is counted as coverage. Contradiction is counted as active hallucination. Neutral is counted as silence.

\paragraph{Nugget verification.}
Claim-level factuality is evaluated with Gemini~2.5~Flash \citep{geminiteam2025gemini25} at temperature~0. Stage~1 extracts atomic claims using \texttt{max\_tokens}=1{,}024. Stage~2 verifies each claim against the relevant catalogue value using \texttt{max\_tokens}=200. Prompts are shown in Prompts~\ref{prompt:nugget-extraction} and~\ref{prompt:nugget-verification}. Primary fields are shown with ground-truth values, while bonus fields are shown only as field concepts. Malformed JSON responses are retried twice. Persistent failures are marked \texttt{not\_verifiable} and account for fewer than 1\% of calls. Due to API budget constraints, nugget verification is run on 6{,}120 descriptions stratified by region, model, and condition.

\begin{table*}[h]
\centering
\small
\begin{tabular}{p{3.8cm}lll}
\toprule
\textbf{Model} & \textbf{Version} & \textbf{Task} & \textbf{API} \\
\midrule
GPT-4o \citep{Achiam2023GPT4TR}
  & \texttt{gpt-4o-2024-08-06} & Generation & OpenAI \\
Llama-4-Maverick \citep{meta2025llama4}
  & \texttt{Llama-4-Maverick-17B-128E-Instruct-FP8} & Generation & DeepInfra \\
Llama-3.2-Vision \citep{Dubey2024TheL3}
  & \texttt{Llama-3.2-11B-Vision-Instruct} & Generation & DeepInfra \\
Qwen2.5-VL \citep{Bai2025Qwen25VLTR}
  & \texttt{Qwen2.5-VL-32B-Instruct} & Generation & DeepInfra \\
Gemini~2.5~Flash \citep{geminiteam2025gemini25}
  & \texttt{gemini-2.5-flash} & Evaluation & Google \\
\bottomrule
\end{tabular}
\caption{Model versions and APIs. LLaMA-3.2-Vision is restricted to the four base Phase~2 conditions. Gemini~2.5~Flash is used only for nugget extraction and verification.}
\label{tab:model-versions}
\end{table*}

\subsection{Claim-Level Factuality Pipeline Validation}
\label{app:validation-checks}

We run two sanity checks on the nugget pipeline. First, extracted spans must occur verbatim in the generated text. Span validity is 98\% over the 6{,}120 evaluated descriptions. Second, NLI predictions are compared with nugget verdicts for primary fields. Entailment should align with \texttt{supported}, and contradiction should align with \texttt{hallucinated}. Agreement over primary field--generation pairs ($n\approx30{,}000$) exceeds Cohen's $\kappa>0.70$. Most disagreements are cases where NLI returns \texttt{neutral} but Stage~2 finds span-level support.

\begin{table*}[!t]
\centering
\setlength{\tabcolsep}{4pt}
\begin{tabular}{lll}
\toprule
\textbf{Field} & \textbf{Template} & \textbf{Example $h_f$} \\
\midrule
Materials
  & This object is made of [$v$].
  & \textit{\ldots made of bronze.} \\
Culture
  & This object belongs to the [$v$] culture.
  & \textit{\ldots Roman culture.} \\
Production date
  & This object was produced [$v$].
  & \textit{\ldots produced 100 BC.} \\
Find spot
  & This object was found at [$v$].
  & \textit{\ldots found at Pompeii.} \\
Place
  & This object originates from [$v$].
  & \textit{\ldots from Roman Empire.} \\
Associated name
  & This object is associated with [$v$].
  & \textit{\ldots associated with Augustus.} \\
Subjects
  & This object depicts [$v$].
  & \textit{\ldots depicts a battle scene.} \\
Technique
  & This object was produced using [$v$].
  & \textit{\ldots using lost-wax casting.} \\
Inscription
  & This object bears the inscription: [$v$].
  & \textit{\ldots inscription: DEO MARTI.} \\
\bottomrule
\end{tabular}
\caption{NLI hypothesis templates and instantiated examples. [$v$] is substituted with ground-truth value $v_f$.}
\label{tab:nli_templates}
\end{table*}

\begin{table*}[!t]
\centering
\setlength{\tabcolsep}{6pt}
\begin{tabular}{lp{9.5cm}}
\toprule
\textbf{Error type} & \textbf{Definition} \\
\midrule
\texttt{wrong\_material}
  & Incorrect material stated (e.g., \textit{iron} for
    ground-truth \textit{bronze}) \\[2pt]
\texttt{wrong\_date}
  & Incorrect production date, century, or period \\[2pt]
\texttt{wrong\_culture}
  & Incorrect cultural or regional attribution \\[2pt]
\texttt{wrong\_place}
  & Incorrect geographic or archaeological provenance \\[2pt]
\texttt{wrong\_subject}
  & Incorrect iconographic or figurative content \\[2pt]
\texttt{wrong\_technique}
  & Incorrect production or manufacturing technique \\[2pt]
\texttt{fabricated\_detail}
  & Plausible elaboration with no anchor in any metadata
    field \\
\bottomrule
\end{tabular}
\caption{Hallucination error taxonomy. Each hallucinated nugget receives one error type and a severity label.}
\label{tab:error_taxonomy}
\end{table*}

\begin{promptbox}[generation]{Generation Prompt}
\small
\textbf{[System]}\\
You are an expert museum curator writing object labels for a
public gallery. Produce clear, professional descriptions
accessible to a general audience.

\medskip
\textbf{[User]}\\
Write a museum label for the artefact described by the
following metadata.

Your response must:
\begin{itemize}[nosep, leftmargin=1.2em]
  \item consists of continuous prose;
  \item contextualise the object culturally, historically, or
        artistically where the metadata supports it;
  \item not reproduce or reference the field label names;
  \item not introduce any information not grounded in the provided
        metadata;
  \item not use bullet points, headers, or lists.
\end{itemize}

\textbf{Metadata}\\
\texttt{\{FIELD\}: \{VALUE\}}\\
\texttt{\{FIELD\}: \{VALUE\}}\\
\ldots

\medskip
\texttt{\{IMAGE appended for image-enabled conditions\}}
\end{promptbox}

\begin{promptbox}[nugget-extraction]{Stage 1: Nugget Extraction Prompt}
\small
\textbf{[System]}\\
You are a specialist fact-checker for museum artefact
descriptions. Extract every factual claim from the provided
description as atomic nuggets grounded in verbatim text spans.

\medskip
\textbf{[User]}\\
Decompose the following museum description into atomic factual
claims (nuggets). For each nugget, provide:
\begin{itemize}[nosep, leftmargin=1.2em]
  \item \textbf{span}: exact verbatim substring from the
        description, which must appear character-for-character;
  \item \textbf{text}: normalised paraphrase of the claim;
  \item \textbf{field}: the metadata field this claim is
        attributable to. Choose from: \textit{Culture},
        \textit{Production date}, \textit{Materials},
        \textit{Find spot}, \textit{Place},
        \textit{Associated name}, \textit{Subjects},
        \textit{Technique}, or \textit{Inscription}. If the claim
        cannot be attributed to any schema field, assign
        \texttt{extra\_content}.
\end{itemize}

\textbf{Provided fields (ground-truth values shown):}\\
\texttt{\{FIELD\}: \{VALUE\}}\\
\texttt{\{FIELD\}: \{VALUE\}}\\
\ldots

\medskip
\textbf{Withheld field concepts (values not provided):}\\
\texttt{\{FIELD\}, \{FIELD\}, \ldots}

\medskip
\textbf{Description:}\\
\texttt{\{GENERATION\}}

\medskip
Return a JSON array of nugget objects. Each nugget should have the
following form:\\
\texttt{\{"span": "...", "text": "...", "field": "..."\}}
\end{promptbox}

\begin{promptbox}[nugget-verification]{Stage 2: Nugget Verification Prompt}
\small
\textbf{[System]}\\
You are an expert museum curator verifying the factual accuracy
of claims about museum artefacts against authoritative catalogue
records.

\medskip
\textbf{[User]}\\
Verify the following claim against the provided ground-truth
metadata value.

Assign one of the following verdicts:
\begin{itemize}[nosep, leftmargin=1.2em]
  \item \textbf{supported}: the claim is factually consistent
        with the ground-truth value;
  \item \textbf{hallucinated}: the claim contradicts or is
        unsupported by the ground-truth value;
  \item \textbf{not\_verifiable}: the claim cannot be confirmed
        or refuted from the ground-truth value alone.
\end{itemize}

If the verdict is \texttt{hallucinated}, additionally assign:
\begin{itemize}[nosep, leftmargin=1.2em]
  \item \textbf{error\_type}: one of
        \texttt{wrong\_material}, \texttt{wrong\_date},
        \texttt{wrong\_culture}, \texttt{wrong\_place},
        \texttt{wrong\_subject}, \texttt{wrong\_technique}, or
        \texttt{fabricated\_detail};
  \item \textbf{severity}: \texttt{minor}, for a factual deviation
        with limited impact, or \texttt{major}, for an error that
        would materially mislead a museum visitor.
\end{itemize}

\textbf{Field:} \texttt{\{FIELD\}}\\
\textbf{Ground-truth value:} \texttt{\{VALUE\}}\\
\textbf{Claim (verbatim span):} \texttt{\{SPAN\}}\\
\textbf{Claim (normalised):} \texttt{\{TEXT\}}

\medskip
Return JSON in the following format:\\
\texttt{\{"verdict": "...", "error\_type": "...",}
\texttt{"severity": "...", "reasoning": "max 20 words"\}}
\end{promptbox}

\section{Additional Results Details}
\label{app:results}

\subsection{Phase 1 Results}
\label{app:phase1-results}

\paragraph{Overall performance.}
\label{app:overall}

\Cref{tab:overall_scores} reports the overall automatic evaluation scores. LLaMA-Maverick achieves the highest mean scores across all reported metrics. Qwen2.5-VL-32B performs worst overall. Differences between LLaMA-Maverick and the remaining models are significant under paired Wilcoxon signed-rank tests computed at the object level. All comparisons have $p<0.05$ and are marked with $\dagger$ in \Cref{tab:overall_scores}.

\begin{table*}[t]
\centering
\setlength{\tabcolsep}{4pt}
\begin{tabular}{lcccc}
\toprule
\textbf{Model} & \textbf{BLEU} & \textbf{ROUGE-L}
  & \textbf{METEOR} & \textbf{\bertscore{}} \\
\midrule
GPT-4o               & 0.240$^\dagger$ & 0.159$^\dagger$ & 0.353$^\dagger$ & 0.856$^\dagger$ \\
LLaMA-Maverick       & \textbf{0.309}  & \textbf{0.196}  & \textbf{0.389}  & \textbf{0.860}  \\
LLaMA-3.2-Vision     & 0.229$^\dagger$ & 0.159$^\dagger$ & 0.334$^\dagger$ & 0.848$^\dagger$ \\
Qwen2.5-32B          & 0.152$^\dagger$ & 0.105$^\dagger$ & 0.277$^\dagger$ & 0.841$^\dagger$ \\
\bottomrule
\end{tabular}
\caption{Phase~1 surface metrics ecaluation of generated descriptions. Bold = best score. $\dagger$ = significantly worse than best model under paired bootstrap and Wilcoxon tests ($p{<}0.05$).}
\label{tab:overall_scores}
\end{table*}

\paragraph{Place predicts score.}
\label{app:place_significance}
We test whether metric scores vary by geographic origin using Kruskal--Wallis tests, with effect sizes reported as $\eta^2$. Place has a significant effect for every model--metric combination ($p<0.001$ throughout). The effect is largest for \bertscore{} ($\eta^2=0.33$--$0.43$), indicating that place explains up to 43\% of score variance. Effects are smaller for METEOR ($\eta^2\approx0.19$--$0.21$), ROUGE-L ($\eta^2\approx0.10$--$0.15$), and BLEU ($\eta^2\approx0.07$--$0.11$).

\Cref{tab:bertscore_by_place} shows per-place \bertscore{} scores. We measure agreement in regional rankings across models using Kendall's rank concordance coefficient ($W$). \bertscore{} shows near-perfect agreement across models, with $W=0.990$, $\chi^2_9=44.6$, and $p<0.0001$. This indicates that models rank places similarly by surface similarity. METEOR also shows high agreement, with $W=0.877$ and $p<0.0001$. ROUGE-L shows moderate agreement, with $W=0.691$ and $p<0.01$. BLEU shows weak and non-significant agreement, with $W=0.310$ and $p=0.125$. This suggests that BLEU place rankings are inconsistent across models, consistent with BLEU's higher sensitivity to output length variation.

\begin{table*}[t!]
\centering
\setlength{\tabcolsep}{5pt}

\begin{tabular}{lcccc c}
\toprule
\textbf{Place}
& \textbf{GPT-4o}
& \textbf{LLaMA-3.2-Vision}
& \textbf{LLaMA-Maverick}
& \textbf{Qwen2.5-32B}
& \textbf{Mean} \\
\midrule

British Isles   & 0.862 & 0.850 & \textbf{0.867} & \cellcolor{gray!15}0.847 & 0.857 \\
Central America & 0.872 & 0.857 & \textbf{0.875} & \cellcolor{gray!15}0.858 & \textbf{0.866} \\
China           & 0.870 & 0.855 & \textbf{0.872} & \cellcolor{gray!15}0.853 & 0.863 \\
Egypt           & 0.850 & 0.845 & \textbf{0.855} & \cellcolor{gray!15}0.837 & 0.847 \\
Greece          & 0.846 & 0.842 & \textbf{0.851} & \cellcolor{gray!15}0.831 & \cellcolor{gray!15}0.843 \\
India           & 0.852 & 0.849 & \textbf{0.859} & \cellcolor{gray!15}0.839 & 0.850 \\
Iran            & 0.859 & 0.850 & \textbf{0.862} & \cellcolor{gray!15}0.841 & 0.853 \\
Iraq            & 0.855 & 0.849 & \textbf{0.862} & \cellcolor{gray!15}0.840 & 0.852 \\
Japan           & 0.865 & 0.853 & \textbf{0.868} & \cellcolor{gray!15}0.850 & 0.859 \\
Roman Empire    & 0.844 & 0.844 & \textbf{0.850} & \cellcolor{gray!15}0.833 & 0.843 \\

\bottomrule
\end{tabular}

\caption{\bertscore{} by place and model. Bold indicates the best model per place; gray shading indicates the lowest score per place. The mean column summarizes overall difficulty per region. Differences across places are significant for all models ($p < 0.001$, Kruskal--Wallis and permutation tests).}

\label{tab:bertscore_by_place}
\end{table*}

\paragraph{Field count.}
\Cref{tab:fieldcount_bertscore_by_place} and \Cref{tab:regional_effects} show that the relationship between metadata quantity and \bertscore{} differs by region. For Greece ($\rho=-0.350$), India ($\rho=-0.411$), and Iran ($\rho=-0.337$), higher field counts are associated with lower \bertscore{} within the same region. These regions also show the largest increases in output length per additional field, ranging from 14.9 to 21.0 words per field. This suggests that richer metadata may induce longer or more specialised descriptions that are less similar to the reference under surface metrics. Roman Empire is the main exception ($\rho=0.040$). In this region, \bertscore{} increases from 0.840 at six fields to 0.847 at nine fields. Roman Empire is also the only region where output length does not increase with field count ($\Delta=-0.8$ words per field). This is consistent with its positive \bertscore{}--field-count relationship and suggests that additional metadata is more closely aligned with reference wording in this region.

\begin{table*}[t]
\centering
\setlength{\tabcolsep}{4pt}
\begin{tabular}{lccccccc}
\toprule
\textbf{Region} & \textbf{3} & \textbf{4} & \textbf{5} & \textbf{6}
& \textbf{7} & \textbf{8} & \textbf{9} \\
\midrule
British Isles$^{*}$         & 0.863 & \textbf{0.864} & 0.857 & 0.856 & 0.852 & 0.854 & -- \\
Central America$^{\dagger}$ & -- & 0.865 & 0.865 & \textbf{0.865} & 0.858 & -- & -- \\
China$^{\dagger}$            & -- & -- & -- & \textbf{0.863} & 0.860 & 0.858 & -- \\
Egypt$^{\dagger}$            & -- & -- & -- & -- & \textbf{0.846} & 0.846 & 0.841 \\
Greece$^{*}$                 & -- & -- & \textbf{0.853} & 0.851 & 0.845 & 0.840 & 0.835 \\
India$^{*}$                  & -- & 0.861 & \textbf{0.868} & 0.862 & 0.847 & 0.845 & -- \\
Iran$^{*}$                   & 0.858 & \textbf{0.859} & 0.856 & 0.856 & 0.849 & 0.847 & 0.850 \\
Iraq$^{*}$                   & -- & -- & \textbf{0.859} & 0.852 & 0.852 & 0.848 & 0.848 \\
Japan$^{*}$                  & 0.863 & 0.860 & \textbf{0.863} & 0.858 & 0.852 & 0.849 & -- \\
Roman Empire                 & -- & -- & -- & 0.840 & 0.842 & 0.843 & \textbf{0.847} \\
\bottomrule
\end{tabular}
\caption{\bertscore{} by region and number of available metadata fields,
averaged across all models. Bold indicates the highest score per region.
$^{*}$~scores differ significantly across field counts for all models
(Kruskal--Wallis, $p<0.05$). $^{\dagger}$~significant for at least one
model. Roman Empire shows no significant variation across field counts.}
\label{tab:fieldcount_bertscore_by_place}
\end{table*}

\begin{table}[!t]
\centering
\small
\begin{tabular}{lll}
\toprule
\textbf{Region}
  & \textbf{Spearman's $\rho$}
  & \textbf{Slope} \\
\midrule
India           & $-0.411$ & 21.0 \\
Greece          & $-0.350$ & 20.6 \\
Iran            & $-0.337$ & 14.9 \\
Egypt           & $-0.089$ & 14.1 \\
Iraq            & $-0.155$ & 13.8 \\
British Isles   & $-0.164$ &  8.8 \\
Japan           & $-0.295$ &  8.3 \\
Central America & $-0.031^{\dagger}$ &  6.3 \\
China           & $-0.092$ &  2.9 \\
Roman Empire    & $\phantom{-}0.040^{\dagger}$ & $-0.8$ \\
\bottomrule
\end{tabular}
\caption{Within-region relationships between metadata field count and
\bertscore{} in Phase~1. The correlation column reports Spearman's
$\rho$ between field count and \bertscore{}. The slope column reports
the average increase in output length, measured as additional words per
metadata field. $^{\dagger}$~not significant at $p<0.05$; all other
correlations are significant.}
\label{tab:regional_effects}
\end{table}

\subsection{Phase~2 Results}
\label{app:phase2-results}

\paragraph{Surface metrics performance}
\label{app:place-surface}

Table~\ref{tab:phase2_model_avg_surface_metrics} reports surface-metric performance for the three models evaluated across all ten Phase~2 conditions; LLaMA-Vision is excluded because it was evaluated on four conditions only and is therefore not directly comparable. GPT-4o achieves the highest \bertscore{} (0.884) and METEOR (0.251), while Qwen2.5-32B achieves the highest BLEU-1 score (0.398). \bertscore{} is compressed across models (range 0.009), consistent with the length confound identified in Phase~1.

\begin{figure*}[!t]
    \centering
    \includegraphics[width=1\linewidth]{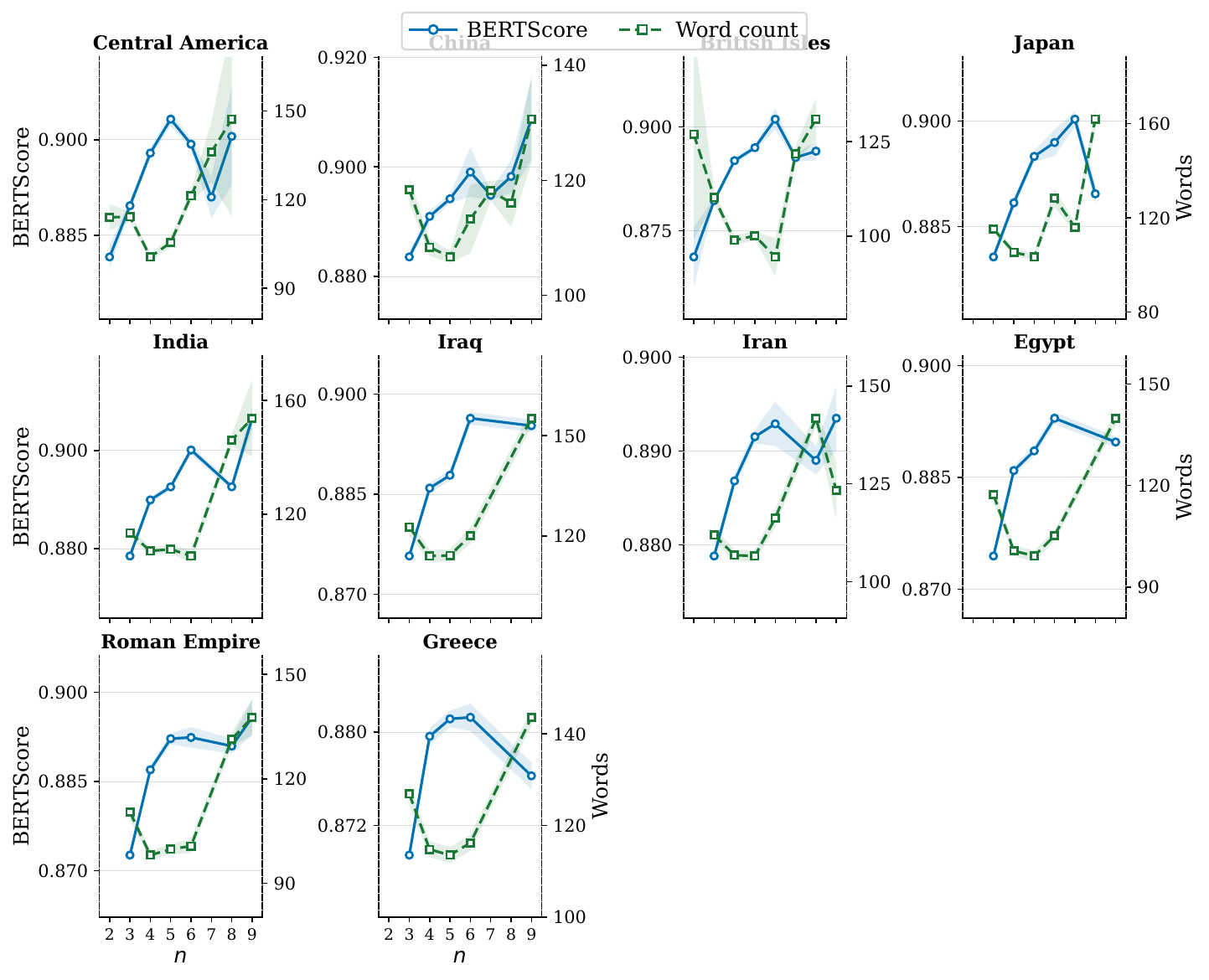}
    \caption{Mean BERTScore and generated word count as a function
of observed metadata field count~$n$ for each of the ten
regions in Phase~2, averaged across all models and
conditions.
}
    \label{fig:bertscore_wc_places}
\end{figure*}

Table~\ref{tab:bertscore_place_model} reports mean \bertscore{} by region and model. Among the three models evaluated across all ten conditions, GPT-4o obtains the highest regional means in most regions, while Qwen2.5-VL-32B is generally lower.
Figure~\ref{fig:bertscore_wc_places} shows per-region field-count--\bertscore{} curves. Peak field counts differ across regions: British Isles peaks early ($n=6$), Japan and Iraq peak at intermediate counts ($n=7$ and $n=6$, respectively), while China and India show broadly increasing curves peaking at $n=9$. Greece peaks at $n=4$--$5$ and declines at higher field counts, consistent with its weak positive field-count correlation ($r=+0.081$) and stronger negative word-count correlation ($r=-0.514$), suggesting that richer metadata is associated with longer outputs that are less similar to the reference under surface metrics~(\Cref{tab:corr_place_nfields_length}. Central America shows a non-monotonic curve with no clear peak. Word-count correlations are uniformly negative and larger in magnitude than field-count correlations across all regions (overall $r=-0.543$ versus $r=+0.201$), with China
($\rho=-0.711$) and Central America ($r=-0.697$) showing the strongest length--\bertscore{} associations, suggesting that output length is a stronger predictor of \bertscore{} variation than metadata quantity alone.

\begin{table*}[!t]
\centering
\small

\begin{tabular}{lrrrr}
\toprule
\textbf{Model}
& \textbf{BLEU-1}
& \textbf{ROUGE-L}
& \textbf{METEOR}
& \textbf{\bertscore{}} \\
\midrule
GPT-4o          & 0.381 & 0.271 & 0.251 & 0.884 \\
LLaMA-Maverick  & 0.327 & 0.244 & 0.213 & 0.877 \\
Qwen2.5-32B     & 0.398 & 0.246 & 0.247 & 0.875 \\
\bottomrule
\end{tabular}
\caption{Phase~2 model-level surface-metric performance averaged across Phase~2 conditions. Values are condition-level means averaged over all probing conditions.}
\label{tab:phase2_model_avg_surface_metrics}
\end{table*}

\begin{table*}[t]
\centering
\small

\begin{tabular}{lccccr}
\toprule
\textbf{region} & \textbf{GPT-4o} & \textbf{LLaMA-Mav.} & \textbf{LLaMA-Vis.} & \textbf{Qwen2.5} & \textbf{Overall} \\
\midrule
\textit{Central America} & \textbf{0.9091} & \textbf{0.9033} & \textbf{0.8769} & 0.8823 & \textbf{0.8957} \\
\textit{China} & 0.9084 & 0.8952 & 0.8762 & 0.8767 & 0.8914 \\
\textit{British Isles} & 0.9001 & 0.8967 & 0.8749 & 0.8821 & 0.8908 \\
\textit{Japan} & 0.8991 & 0.8947 & 0.8761 & 0.8802 & 0.8895 \\
\textit{India} & 0.8964 & 0.8924 & 0.8760 & \textbf{0.8832} & 0.8890 \\
\textit{Iraq} & 0.8978 & 0.8915 & 0.8739 & 0.8800 & 0.8879 \\
\textit{Iran} & 0.8978 & 0.8884 & 0.8724 & 0.8785 & 0.8864 \\
\textit{Egypt} & 0.8936 & 0.8897 & 0.8727 & 0.8794 & 0.8858 \\
\textit{Roman Empire} & 0.8949 & 0.8906 & 0.8716 & 0.8776 & 0.8858 \\
\textit{Greece} & 0.8864 & 0.8809 & 0.8637 & 0.8690 & 0.8770 \\
\bottomrule
\end{tabular}
\caption{Phase~2 mean BERTScore by region and model. Place differences are significant for all models (Kruskal--Wallis, all $p<0.001$; Bold indicates highest value per column. LLaMA-Vision ran on conditions R, S, S\_noi, T only.}
\label{tab:bertscore_place_model}
\end{table*}

\begin{table*}[t]
\centering
\small
\setlength{\tabcolsep}{5pt}

\begin{tabular}{lcc|cc}
\toprule
& \multicolumn{2}{c|}{\textbf{Field count} ($n$)}
& \multicolumn{2}{c}{\textbf{Word count}} \\
\cmidrule(lr){2-3}\cmidrule(l){4-5}
\textbf{region}
  & $r$\rlap{***} & $\rho$\rlap{***}
  & $r$\rlap{***} & $\rho$\rlap{***} \\
\midrule
\textit{Central America} & $+0.248$ & $+0.257$ & $-0.697$ & $-0.704$ \\
\textit{China}           & $+0.216$ & $+0.223$ & $-0.700$ & $\mathbf{-0.711}$ \\
\textit{British Isles}   & $+0.223$ & $+0.263$ & $-0.647$ & $-0.618$ \\
\textit{Japan}           & $+0.320$ & $+0.363$ & $-0.592$ & $-0.578$ \\
\textit{India}           & $+0.359$ & $+0.417$ & $-0.424$ & $-0.398$ \\
\textit{Iraq}            & $\mathbf{+0.381}$ & $+0.421$ & $-0.531$ & $-0.512$ \\
\textit{Iran}            & $+0.182$ & $+0.359$ & $-0.504$ & $-0.479$ \\
\textit{Egypt}           & $+0.320$ & $+0.384$ & $-0.521$ & $-0.486$ \\
\textit{Roman Empire}    & $+0.301$ & $+0.394$ & $-0.421$ & $-0.400$ \\
\textit{Greece}          & $+0.081$ & $+0.200$ & $-0.514$ & $-0.547$ \\
\midrule
\textit{Overall}         & $+0.201$ & $+0.285$ & $\mathbf{-0.543}$ & $-0.544$ \\
\bottomrule
\end{tabular}
\caption{Phase~2 Pearson ($r$) and Spearman ($\rho$) correlations between BERTScore and number of metadata fields~$n$ (left panel) and generated word count (right panel), computed per region across all conditions and models. All correlations are significant at $p < 0.001$ (***). Bold marks the strongest correlation per column. }
\label{tab:corr_place_nfields_length}
\end{table*}

\subsubsection{Per-field Analysis}
\label{app:place-halluc}

\begin{table*}[!t]
\centering\small

\begin{tabular}{lcccccc}
\toprule
& \multicolumn{3}{c}{\textbf{\textsc{S\_ctx} arm}}
& \multicolumn{3}{c}{\textbf{\textsc{S\_cont} arm}} \\
\cmidrule(lr){2-4}\cmidrule(lr){5-7}
\textbf{Field} & Cor & Wro & Sil & Cor & Wro & Sil \\
\midrule
\textit{Technique}   & 10.6          & 51.1          & 38.3          &  9.8          & 56.8          & 33.4          \\
\textit{Subjects}    & 17.9          & 49.7          & 32.4          &  ---          &  ---          &  ---          \\
\textit{Inscription} & \textbf{27.2} & 13.3          & 59.5          & \textbf{10.6} &  9.4          & 80.1          \\
\textit{Assoc name}  &  2.9          & \textbf{4.3}  & \textbf{92.9} &  8.1          &  7.7          & 84.2          \\
\textit{Place}       & 11.2          & 20.2          & 68.5          &  3.7          &  2.5          & 93.8          \\
\textit{Find spot}   &  ---          &  ---          &  ---          &  0.1          & \textbf{0.6}  & \textbf{99.4} \\
\bottomrule
\end{tabular}
\caption{Phase~2 bonus field rates split by arm type: the \textsc{S\_ctx} arm and \textsc{S\_cont} arm .  All values are percentages. Bold indicates best
value per column within each arm: highest for Cor and Sil; lowest for Wro. Fields absent from a arm are shown as~---.}
\label{tab:field_bonus_ladder}
\end{table*}

\begin{figure*}[!t]
    \centering
    \includegraphics[width=0.8\linewidth]{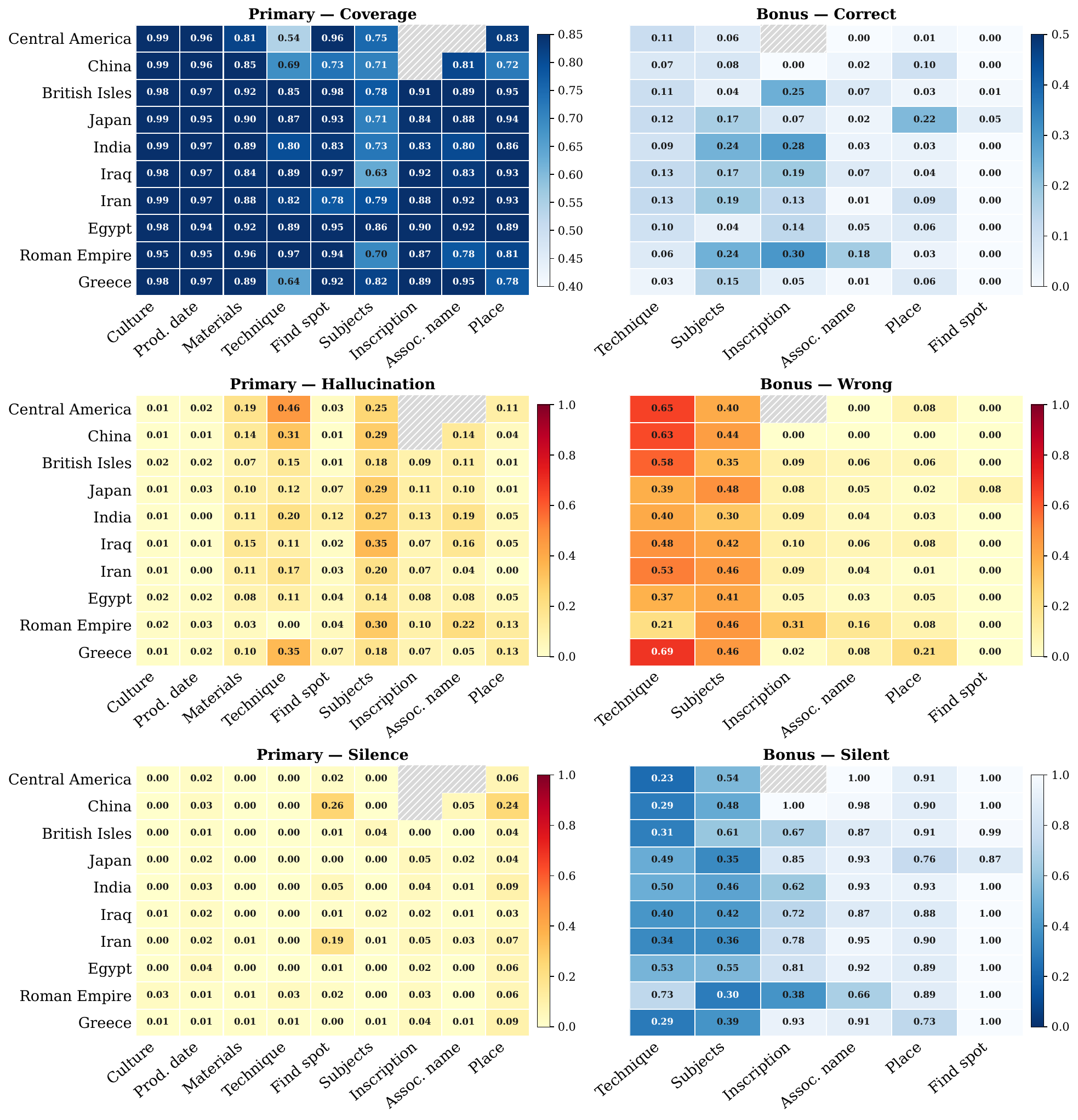}
    \caption{Nugget-verified field outcomes by region and field, averaged across all models and conditions. Each row pairs a primary outcome (field provided, left) with its corresponding bonus outcome (field withheld, right): coverage with bonus correct (top), hallucination with bonus wrong (middle), and silence with bonus silent (bottom). Blue panels indicate outcomes where higher values are better; yellow--red panels indicate outcomes where higher values are worse. Hatched cells indicate fewer than 15 observations.}
    \label{fig:field_place_3x2_full}
\end{figure*}

Figure~\ref{fig:field_place_3x2_full} shows primary coverage and bonus correct panels. Primary coverage is high across most field--region combinations ($> 0.90$), with the lowest values concentrated in \textit{Technique} and \textit{Subjects} for Central America and China. Bonus correct rates are uniformly low, with \textit{Inscription} showing the highest correct recovery across regions, reaching $0.30$ for Roman Empire, $0.28$ for India, and $0.25$ for British Isles, consistent with the aggregate field-level finding that \textit{Inscription} is the only withheld field for which bonus correct exceeds bonus wrong. \textit{Find spot} shows near-zero bonus correct across all regions. \Cref{tab:field_bonus_ladder} shows a per-field breakdown across the two field type ladders.

\section{Qualitative Analysis}
\label{app:qualitative-errors}
\subsection{Qualitative Error Patterns}
We complement the aggregate claim-level analysis with qualitative inspection of hallucinated and silent field outcomes. The dominant pattern is schema-boundary overreach: models generate claims that are plausible in the artefact domain yet unsupported by the specific catalogue field being evaluated. Table~\ref{tab:qualitative_error_patterns} summarises the main error mechanisms.

\begin{table*}[!t]
\centering
\small
\setlength{\tabcolsep}{4pt}
\begin{tabular}{p{0.20\linewidth}p{0.16\linewidth}p{0.34\linewidth}p{0.22\linewidth}}
\toprule
\textbf{Error pattern} &
\textbf{Main fields} &
\textbf{Mechanism} &
\textbf{Example pattern} \\
\midrule

Field substitution &
\textit{Place}, \textit{Find spot} &
The model transfers a value from one metadata field into another, especially using an excavation site, city, or acquisition location as the broader \textit{Place} value. &
\textit{Find spot}: Babylon / Nineveh / Marcianopolis $\rightarrow$ \textit{Place}. \\

Geographic over-specification &
\textit{Place} &
The model narrows a broad catalogue place label into a more specific region, city, or modern location that is not supported by the target field value. &
\textit{Central America} $\rightarrow$ northern Peten / Guatemala; \textit{Roman Empire} $\rightarrow$ Rome. \\

Material-to-object inference &
\textit{Materials} &
The model uses a material label to infer an object type or function, rather than restricting the claim to material composition. &
\textit{Earthenware} $\rightarrow$ vessel; \textit{stone} $\rightarrow$ stele; \textit{gold} $\rightarrow$ coin / jewellery. \\

Field-boundary confusion &
\textit{Materials}, \textit{Technique} &
The model assigns a value to a neighbouring field, producing a plausible claim that violates the catalogue schema. &
\textit{Glaze} or \textit{velvet} treated as material when recorded as technique. \\

Art-historical over-specification &
\textit{Technique} &
The model refines a broad catalogue technique into a more specific art-historical technique using domain priors or visual cues. &
\textit{Painted} $\rightarrow$ red-figure / black-figure. \\

Visual elaboration beyond vocabulary &
\textit{Subjects} &
The model adds visible motifs, attributes, posture, or iconographic details not present in the controlled subject vocabulary. &
\textit{Bird; phoenix} $\rightarrow$ foliage, vines, secondary motifs. \\

Bonus technique confabulation &
\textit{Technique} &
When technique is withheld, the model produces art-critical or stylistic descriptions instead of recovering catalogue technique values. &
\textit{Plated} $\rightarrow$ intricate engraving; \textit{drilled} $\rightarrow$ smooth and polished. \\

Temporal/contextual expansion &
\textit{Production date}, \textit{Place} &
The model adds broad temporal or contextual qualifiers that are plausible but not supported by the metadata field. &
\textit{Egypt} $\rightarrow$ ancient Egypt; date ranges expanded beyond the recorded value. \\

Documentation redundancy and silence &
\textit{Find spot}, \textit{Place} &
The model omits a populated field when its value duplicates another field or adds little distinct information. &
\textit{Find spot}: Found/Acquired: China omitted after mentioning \textit{Place}: China. \\

\bottomrule
\end{tabular}
\caption{Qualitative schema-boundary error patterns identified from hallucinated and silent field-level outcomes. The examples illustrate recurring mechanisms rather than exhaustive error categories.}
\label{tab:qualitative_error_patterns}
\end{table*}

\paragraph{Place and Find spot.}
Errors involving \textit{Place} and \textit{Find spot} arise from substitution, over-specification, or redundancy. Models sometimes assign a \textit{Find spot} value to \textit{Place}; for example, an object catalogued with \textit{Place}: \textit{Central America} is described as \textit{Meso-America}, a value present only in \textit{Find spot}. In other cases, models narrow broad place labels using information from \textit{Culture} or \textit{Production date}, such as describing a Greek object as \textit{ancient Greece}. Models also add temporal or contextual qualifiers not supported by the target field: objects catalogued with broad \textit{Place} values such as \textit{Egypt} or \textit{India} receive descriptors such as \textit{ancient}, and \textit{Production date} ranges are occasionally extended beyond the recorded value. Conversely, some apparent \textit{Find spot} silence reflects redundant documentation rather than omission: several Chinese records use values such as \textit{Found/Acquired: China}, which add no geographic information beyond the broader \textit{Place} value. These cases illustrate how geographic claims may remain semantically plausible while violating the intended distinction between neighbouring catalogue fields.

\paragraph{Technique.}
Technique errors reflect over-specification when the field is present and unsupported curatorial elaboration when it is withheld. When \textit{Technique} is provided, models sometimes replace broad catalogue values with more specialised art-historical subtypes; Greek objects catalogued simply as \textit{painted} are described using terms such as \textit{black-figure technique}. When \textit{Technique} is withheld, models generate aesthetic or interpretive descriptions rather than recoverable manufacturing processes: objects with withheld values such as \textit{painted}, \textit{carved}, or \textit{gilded} receive claims about naturalism, craftsmanship, or calligraphic style. These outputs are domain-plausible but not supported by the metadata field being evaluated, explaining the high bonus wrong rates for \textit{Technique}: absent process information invites stylistic elaboration rather than recovery of catalogue-vocabulary manufacturing terms.

\paragraph{Subjects.}
For \textit{Subjects}, models frequently elaborate beyond the controlled subject label through visual and compositional inference. Controlled-vocabulary subject labels are often sparse relative to the visual complexity of the depicted object; models expand the label using visual priors and domain knowledge drawn from pretraining, generating content that is image-consistent but not catalogue-consistent. Records labelled only as \textit{bird} may receive additional claims about foliage, vine motifs, or decorative scenery, while a record labelled \textit{horse/ass} may be described as including a detailed saddle. The resulting claims are not supported by the catalogue subject value but reflect legitimate visual observations of content not captured by the controlled vocabulary.

\paragraph{Materials.}
Material mentions are comparatively well grounded, consistent with the low primary hallucination rates for \textit{Materials} in the quantitative results; errors here are edge cases rather than a dominant failure mode. Most arise not from incorrect material identification but from unsupported inference triggered by the material value: \textit{pottery} may be realised as \textit{bowl}, and \textit{jadeite} as \textit{high-quality jadeite}. A related pattern is field-boundary confusion: models occasionally assign a value to a neighbouring field, treating \textit{glaze} or \textit{velvet} as a material when the catalogue records them under \textit{Technique}. These cases show how correctly grounded material mentions can induce additional assumptions about object type or quality, and how schema proximity between fields creates an additional source of boundary error.

The dominant failure mode is therefore not arbitrary hallucination, but boundary overreach across neighbouring schema fields, visual priors, and domain-plausible curatorial conventions. This explains why fluent descriptions can appear plausible to a general reader while violating the intended relation between catalogue field, field value, and generated claim.

\subsection{Generation Samples}
\begin{figure*}[p]
    \centering
    \includegraphics[width=\textwidth, height=0.82\textheight,
                     keepaspectratio]
                    {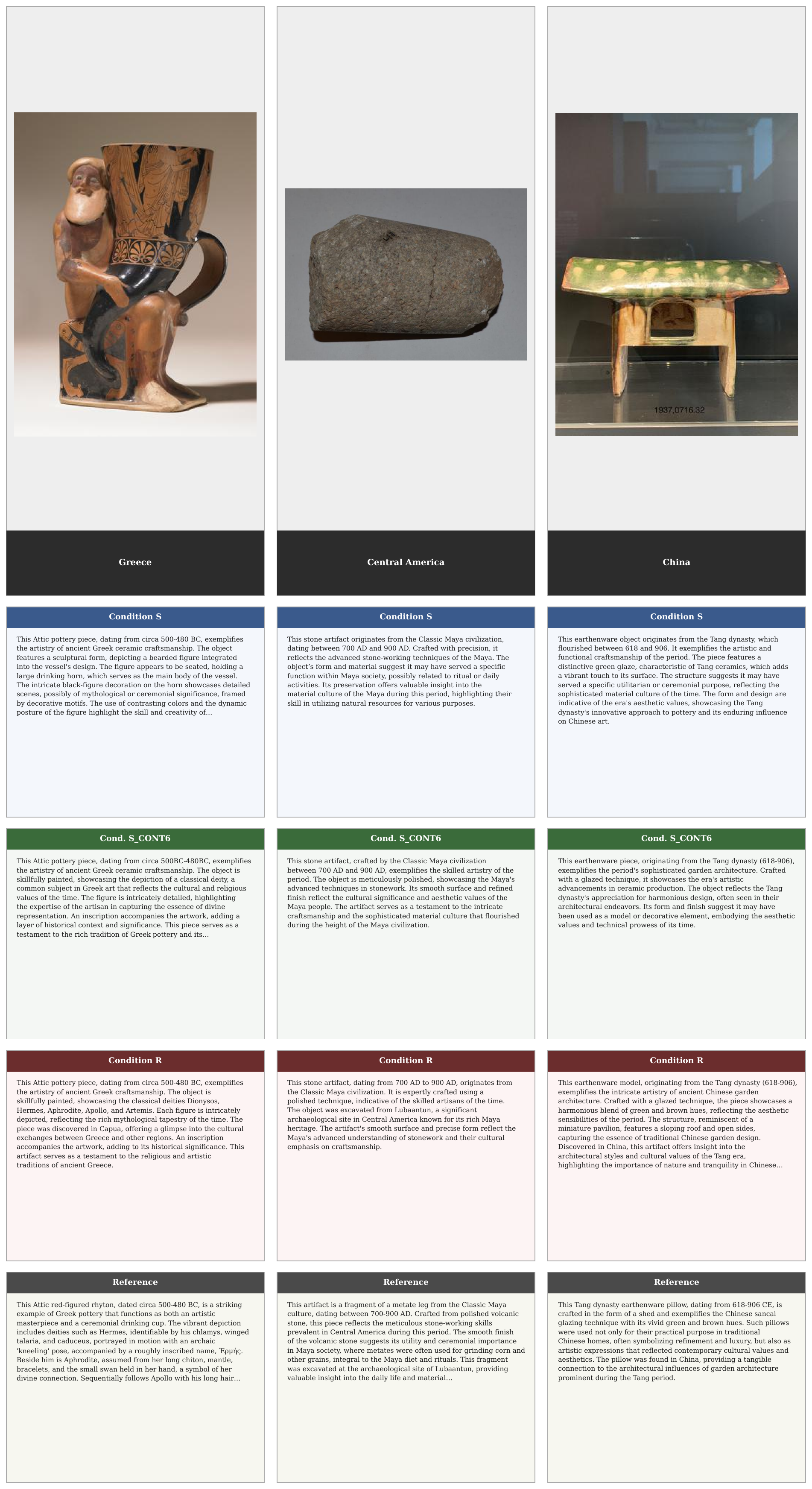}
    \caption{Generated descriptions for three objects across conditions
    \textsc{S} (sparse floor), \textsc{S\_CONT6} (sparse floor plus
    \textit{Subjects}, \textit{Technique}, \textit{Inscription}), and
    \textsc{R} (all available fields with image), alongside the museum
    reference description. Each row shows the same object; descriptions
    are truncated to 100 words.}
    \label{fig:generation_intra_object}
\end{figure*}

\begin{figure*}[p]
    \centering
    \includegraphics[width=\textwidth, height=0.82\textheight,
                     keepaspectratio]
                    {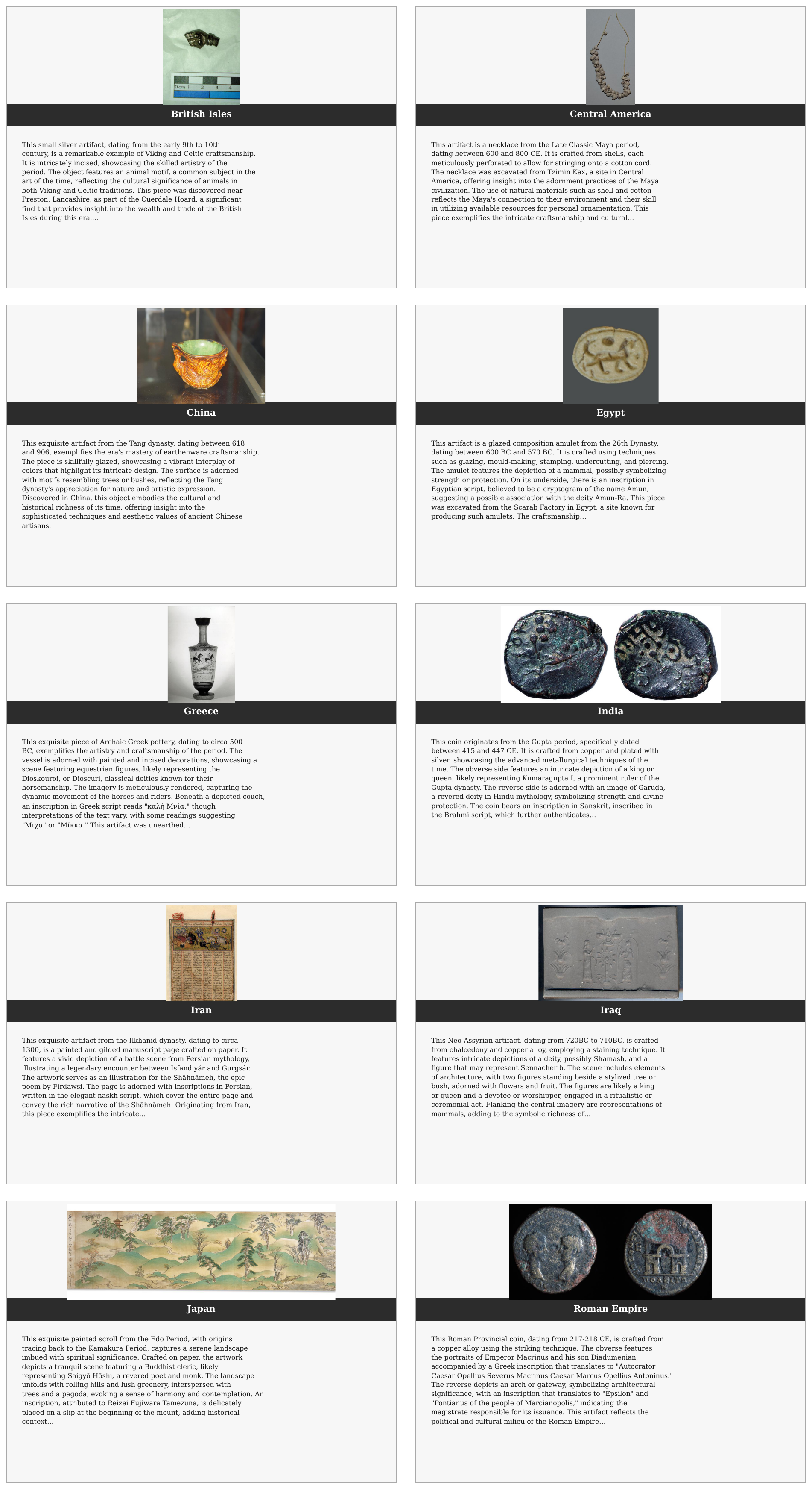}
    \caption{Representative GPT-4o generated descriptions under condition
    \textsc{R} (all available fields with image) for one object per
    civilisation. Each cell shows the museum object and the generated
    description.}
    \label{fig:generation_per_region}
\end{figure*}

\Cref{fig:generation_intra_object} and \Cref{fig:generation_per_region} shows generations samples from GPT4o across condtions and regions.

%% file: main.bbl
\begin{thebibliography}{39}
\providecommand{\natexlab}[1]{#1}

\bibitem[{Arnold and Tilton(2024)}]{arnold-tilton-2024-explainable}
Taylor~B. Arnold and Lauren Tilton. 2024.
\newblock \href {https://api.semanticscholar.org/CorpusID:273877915}
  {Explainable search and discovery of visual cultural heritage collections
  with multimodal large language models}.
\newblock In \emph{Workshop on Computational Humanities Research}.

\bibitem[{Bai et~al.(2025)Bai, Chen, Liu, Wang, Ge, Song, Dang, Wang, Wang,
  Tang, Zhong, Zhu, Yang, Li, Wan, Wang, Ding, Fu, Xu, Ye, Zhang, Xie, Cheng,
  Zhang, Yang, Xu, and Lin}]{Bai2025Qwen25VLTR}
Shuai Bai, Keqin Chen, Xuejing Liu, Jialin Wang, Wenbin Ge, Sibo Song, Kai
  Dang, Peng Wang, Shijie Wang, Jun Tang, Humen Zhong, Yuanzhi Zhu, Mingkun
  Yang, Zhaohai Li, Jianqiang Wan, Pengfei Wang, Wei Ding, Zheren Fu, Yiheng
  Xu, and 8 others. 2025.
\newblock \href {https://api.semanticscholar.org/CorpusID:276449796}
  {{Qwen2.5-VL} technical report}.
\newblock \emph{ArXiv}, abs/2502.13923.

\bibitem[{Banerjee and Lavie(2005)}]{banerjee-lavie-2005-meteor}
Satanjeev Banerjee and Alon Lavie. 2005.
\newblock \href {https://aclanthology.org/W05-0909/} {{METEOR}: An automatic
  metric for {MT} evaluation with improved correlation with human judgments}.
\newblock In \emph{Proceedings of the {ACL} Workshop on Intrinsic and Extrinsic
  Evaluation Measures for Machine Translation and/or Summarization}, pages
  65--72, Ann Arbor, Michigan. Association for Computational Linguistics.

\bibitem[{Battaner and Spence(2026)}]{Battaner-2026}
Elena Battaner and Paul Spence. 2026.
\newblock \href {https://doi.org/10.5334/johd.474} {The linguistic asymmetry
  index ({LAI}): Benchmarking equity in multilingual research infrastructures}.
\newblock \emph{Journal of Open Humanities Data}.

\bibitem[{Bruce and Hillmann(2004)}]{bruce2004continuum}
Thomas~R. Bruce and Diane~I. Hillmann. 2004.
\newblock The continuum of metadata quality: Defining, expressing, exploiting.
\newblock In Diane~I. Hillmann and Elaine~L. Westbrooks, editors,
  \emph{Metadata in Practice}, pages 238--256. ALA Editions, Chicago.

\bibitem[{Bu et~al.(2025)Bu, Wang, Wang, and Liu}]{bu-etal-2025-investigation}
Fan Bu, Zheng Wang, Siyi Wang, and Ziyao Liu. 2025.
\newblock \href {https://api.semanticscholar.org/CorpusID:275336220} {An
  investigation into value misalignment in {LLM}-generated texts for cultural
  heritage}.
\newblock \emph{ArXiv}, abs/2501.02039.

\bibitem[{Candela et~al.(2023)Candela, Pereda, S\'{a}ez, Escobar, S\'{a}nchez,
  Torres, Palacios, McDonough, and
  Murrieta-Flores}]{candela-etal-2023-colonial}
Gustavo Candela, Javier Pereda, Dolores S\'{a}ez, Pilar Escobar, Alexander
  S\'{a}nchez, Andr\'{e}s~Villa Torres, Albert~A. Palacios, Kelly McDonough,
  and Patricia Murrieta-Flores. 2023.
\newblock \href {https://doi.org/10.1145/3594727} {An ontological approach for
  unlocking the colonial archive}.
\newblock \emph{J. Comput. Cult. Herit.}, 16(4).

\bibitem[{Du{\v{s}}ek and Kasner(2020)}]{dusek-kasner-2020-evaluating}
Ond{\v{r}}ej Du{\v{s}}ek and Zden{\v{e}}k Kasner. 2020.
\newblock \href {https://doi.org/10.18653/v1/2020.inlg-1.19} {Evaluating
  semantic accuracy of data-to-text generation with natural language
  inference}.
\newblock In \emph{Proceedings of the 13th International Conference on Natural
  Language Generation}, pages 131--137, Dublin, Ireland. Association for
  Computational Linguistics.

\bibitem[{Falke et~al.(2019)Falke, Ribeiro, Utama, Dagan, and
  Gurevych}]{falke-etal-2019-ranking}
Tobias Falke, Leonardo F.~R. Ribeiro, Prasetya~Ajie Utama, Ido Dagan, and Iryna
  Gurevych. 2019.
\newblock \href {https://doi.org/10.18653/v1/P19-1213} {Ranking generated
  summaries by correctness: An interesting but challenging application for
  natural language inference}.
\newblock In \emph{Proceedings of the 57th Annual Meeting of the Association
  for Computational Linguistics}, pages 2214--2220, Florence, Italy.
  Association for Computational Linguistics.

\bibitem[{Gardent et~al.(2017)Gardent, Shimorina, Narayan, and
  Perez-Beltrachini}]{gardent-etal-2017-webnlg}
Claire Gardent, Anastasia Shimorina, Shashi Narayan, and Laura
  Perez-Beltrachini. 2017.
\newblock \href {https://doi.org/10.18653/v1/W17-3518} {The {W}eb{NLG}
  challenge: Generating text from {RDF} data}.
\newblock In \emph{Proceedings of the 10th International Conference on Natural
  Language Generation}, pages 124--133, Santiago de Compostela, Spain.
  Association for Computational Linguistics.

\bibitem[{Gemini~Team(2025)}]{geminiteam2025gemini25}
Google Gemini~Team. 2025.
\newblock \href {https://api.semanticscholar.org/CorpusID:280151524} {Gemini
  2.5: Pushing the frontier with advanced reasoning, multimodality, long
  context, and next generation agentic capabilities}.
\newblock \emph{ArXiv}, abs/2507.06261.

\bibitem[{Ghaboura et~al.(2025)Ghaboura, More, Thawkar, Ghallabi, Thawakar,
  Khan, Cholakkal, Khan, and Anwer}]{ghaboura-etal-2025-time}
Sara Ghaboura, Ketan~Pravin More, Ritesh Thawkar, Wafa~Al Ghallabi, Omkar
  Thawakar, Fahad~Shahbaz Khan, Hisham Cholakkal, Salman Khan, and Rao~Muhammad
  Anwer. 2025.
\newblock \href {https://doi.org/10.18653/v1/2025.findings-acl.1211} {Time
  travel: A comprehensive benchmark to evaluate {LMM}s on historical and
  cultural artifacts}.
\newblock In \emph{Findings of the Association for Computational Linguistics:
  ACL 2025}, pages 23627--23641, Vienna, Austria. Association for Computational
  Linguistics.

\bibitem[{Gilliland(2016)}]{gilliland2016setting}
Anne~J. Gilliland. 2016.
\newblock \href
  {https://www.getty.edu/publications/intrometadata/setting-the-stage/}
  {Setting the stage}.
\newblock In Murtha Baca, editor, \emph{Introduction to Metadata}, 3rd edition.
  Getty Research Institute, Los Angeles.

\bibitem[{He et~al.(2021)He, Liu, Gao, and Chen}]{he2021deberta}
Pengcheng He, Xiaodong Liu, Jianfeng Gao, and Weizhu Chen. 2021.
\newblock \href {https://openreview.net/forum?id=XPZIaotutsD} {{DeBERTa}:
  Decoding-enhanced {BERT} with disentangled attention}.
\newblock In \emph{9th International Conference on Learning Representations,
  {ICLR} 2021, Virtual Event, Austria, May 3-7, 2021}. OpenReview.net.

\bibitem[{Hyv\"{o}nen(2012)}]{hyvonen2012publishing}
Eero Hyv\"{o}nen. 2012.
\newblock \href {https://doi.org/10.2200/S00452ED1V01Y201210WBE003}
  {\emph{Publishing and Using Cultural Heritage Linked Data on the Semantic
  Web}}.
\newblock Synthesis Lectures on the Semantic Web: Theory and Technology. Morgan
  \& Claypool.

\bibitem[{Isard et~al.(2018)Isard, Oberlander, and
  Grover}]{isard-etal-2018-cycling}
Amy Isard, Jon Oberlander, and Claire Grover. 2018.
\newblock \href {https://aclanthology.org/L18-1483/} {Up-cycling data for
  natural language generation}.
\newblock In \emph{Proceedings of the Eleventh International Conference on
  Language Resources and Evaluation ({LREC} 2018)}, Miyazaki, Japan. European
  Language Resources Association (ELRA).

\bibitem[{Ji et~al.(2023)Ji, Lee, Frieske, Yu, Su, Xu, Ishii, Bang, Madotto,
  and Fung}]{ji-etal-2023-survey}
Ziwei Ji, Nayeon Lee, Rita Frieske, Tiezheng Yu, Dan Su, Yan Xu, Etsuko Ishii,
  Ye~Jin Bang, Andrea Madotto, and Pascale Fung. 2023.
\newblock \href {https://doi.org/10.1145/3571730} {Survey of hallucination in
  natural language generation}.
\newblock \emph{ACM Comput. Surv.}, 55(12).

\bibitem[{Lewis et~al.(2025)Lewis, Oksanen, Ehrnsten, Rantala, Tuominen, and
  Hyv\"{o}nen}]{lewis-hyvonen2025}
Michael Lewis, Eljas Oksanen, Frida Ehrnsten, Heikki Rantala, Jouni Tuominen,
  and Eero Hyv\"{o}nen. 2025.
\newblock \href {https://doi.org/10.1145/3736770} {The impact of human
  decision-making on the research value of archaeological data}.
\newblock \emph{J. Comput. Cult. Herit.}, 18(3).

\bibitem[{Lin(2004)}]{lin-2004-rouge}
Chin-Yew Lin. 2004.
\newblock \href {https://aclanthology.org/W04-1013/} {{ROUGE}: A package for
  automatic evaluation of summaries}.
\newblock In \emph{Text Summarization Branches Out}, pages 74--81, Barcelona,
  Spain. Association for Computational Linguistics.

\bibitem[{Liu et~al.(2023)Liu, Li, Wu, and Lee}]{liu2024visual}
Haotian Liu, Chunyuan Li, Qingyang Wu, and Yong~Jae Lee. 2023.
\newblock Visual instruction tuning.
\newblock In \emph{Proceedings of the 37th International Conference on Neural
  Information Processing Systems}, NIPS '23, Red Hook, NY, USA. Curran
  Associates Inc.

\bibitem[{{Llama Team, AI @ Meta}(2024)}]{Dubey2024TheL3}
{Llama Team, AI @ Meta}. 2024.
\newblock \href {https://api.semanticscholar.org/CorpusID:271571434} {The
  {Llama} 3 herd of models}.
\newblock \emph{ArXiv}, abs/2407.21783.

\bibitem[{Lorenzini et~al.(2021)Lorenzini, Rospocher, and
  Tonelli}]{LorenziniRospocherTonelli2021}
Matteo Lorenzini, Marco Rospocher, and Sara Tonelli. 2021.
\newblock \href {https://doi.org/10.1093/llc/fqab036} {On assessing metadata
  completeness in digital cultural heritage repositories}.
\newblock \emph{Digital Scholarship in the Humanities}, 36(Supplement
  2):ii182--ii188.

\bibitem[{Matoni et~al.(2026)Matoni, Kesper, and
  Taentzer}]{matoni2026qualitydescriptiveinformationcultural}
Markus Matoni, Arno Kesper, and Gabriele Taentzer. 2026.
\newblock \href {https://arxiv.org/abs/2602.21249} {Quality of descriptive
  information on cultural heritage objects: Definition and empirical
  evaluation}.
\newblock \emph{Preprint}, arXiv:2602.21249.

\bibitem[{Maynez et~al.(2020)Maynez, Narayan, Bohnet, and
  McDonald}]{maynez-etal-2020-faithfulness}
Joshua Maynez, Shashi Narayan, Bernd Bohnet, and Ryan McDonald. 2020.
\newblock \href {https://doi.org/10.18653/v1/2020.acl-main.173} {On
  faithfulness and factuality in abstractive summarization}.
\newblock In \emph{Proceedings of the 58th Annual Meeting of the Association
  for Computational Linguistics}, pages 1906--1919, Online. Association for
  Computational Linguistics.

\bibitem[{{Meta}(2025)}]{meta2025llama4}
{Meta}. 2025.
\newblock \href
  {https://www.llama.com/docs/model-cards-and-prompt-formats/llama4/} {Llama 4
  model cards and prompt formats}.

\bibitem[{Min et~al.(2023)Min, Krishna, Lyu, Lewis, Yih, Koh, Iyyer,
  Zettlemoyer, and Hajishirzi}]{min-etal-2023-factscore}
Sewon Min, Kalpesh Krishna, Xinxi Lyu, Mike Lewis, Wen-tau Yih, Pang Koh, Mohit
  Iyyer, Luke Zettlemoyer, and Hannaneh Hajishirzi. 2023.
\newblock \href {https://doi.org/10.18653/v1/2023.emnlp-main.741}
  {{FA}ct{S}core: Fine-grained atomic evaluation of factual precision in long
  form text generation}.
\newblock In \emph{Proceedings of the 2023 Conference on Empirical Methods in
  Natural Language Processing}, pages 12076--12100, Singapore. Association for
  Computational Linguistics.

\bibitem[{Novikova et~al.(2017)Novikova, Du{\v{s}}ek, and
  Rieser}]{novikova-etal-2017-e2e}
Jekaterina Novikova, Ond{\v{r}}ej Du{\v{s}}ek, and Verena Rieser. 2017.
\newblock \href {https://doi.org/10.18653/v1/W17-5525} {The {E}2{E} dataset:
  New challenges for end-to-end generation}.
\newblock In \emph{Proceedings of the 18th Annual {SIG}dial Meeting on
  Discourse and Dialogue}, pages 201--206, Saarbr{\"u}cken, Germany.
  Association for Computational Linguistics.

\bibitem[{{OpenAI}(2023)}]{Achiam2023GPT4TR}
{OpenAI}. 2023.
\newblock \href {https://arxiv.org/abs/2303.08774} {{GPT-4} technical report}.
\newblock \emph{ArXiv}, abs/2303.08774.

\bibitem[{Pagnoni et~al.(2021)Pagnoni, Balachandran, and
  Tsvetkov}]{pagnoni-etal-2021-understanding}
Artidoro Pagnoni, Vidhisha Balachandran, and Yulia Tsvetkov. 2021.
\newblock \href {https://doi.org/10.18653/v1/2021.naacl-main.383}
  {Understanding factuality in abstractive summarization with {FRANK}: A
  benchmark for factuality metrics}.
\newblock In \emph{Proceedings of the 2021 Conference of the North American
  Chapter of the Association for Computational Linguistics: Human Language
  Technologies}, pages 4812--4829, Online. Association for Computational
  Linguistics.

\bibitem[{Papineni et~al.(2002)Papineni, Roukos, Ward, and
  Zhu}]{papineni-etal-2002-bleu}
Kishore Papineni, Salim Roukos, Todd Ward, and Wei-Jing Zhu. 2002.
\newblock \href {https://doi.org/10.3115/1073083.1073135} {{B}leu: a method for
  automatic evaluation of machine translation}.
\newblock In \emph{Proceedings of the 40th Annual Meeting of the Association
  for Computational Linguistics}, pages 311--318, Philadelphia, Pennsylvania,
  USA. Association for Computational Linguistics.

\bibitem[{Parikh et~al.(2020)Parikh, Wang, Gehrmann, Faruqui, Dhingra, Yang,
  and Das}]{parikh-etal-2020-totto}
Ankur Parikh, Xuezhi Wang, Sebastian Gehrmann, Manaal Faruqui, Bhuwan Dhingra,
  Diyi Yang, and Dipanjan Das. 2020.
\newblock \href {https://doi.org/10.18653/v1/2020.emnlp-main.89} {{ToTTo}: A
  controlled table-to-text generation dataset}.
\newblock In \emph{Proceedings of the 2020 Conference on Empirical Methods in
  Natural Language Processing (EMNLP)}, pages 1173--1186, Online. Association
  for Computational Linguistics.

\bibitem[{Reusens et~al.(2025)Reusens, Adams, and Baesens}]{reusens2025large}
Manon Reusens, Amy Adams, and Bart Baesens. 2025.
\newblock \href {https://doi.org/10.1007/s00146-025-02227-8} {Large language
  models to make museum archive collections more accessible}.
\newblock \emph{AI Soc.}, 40(6):4485–4497.

\bibitem[{Rohrbach et~al.(2018)Rohrbach, Hendricks, Burns, Darrell, and
  Saenko}]{rohrbach-etal-2018-object}
Anna Rohrbach, Lisa~Anne Hendricks, Kaylee Burns, Trevor Darrell, and Kate
  Saenko. 2018.
\newblock \href {https://doi.org/10.18653/v1/D18-1437} {Object hallucination in
  image captioning}.
\newblock In \emph{Proceedings of the 2018 Conference on Empirical Methods in
  Natural Language Processing}, pages 4035--4045, Brussels, Belgium.
  Association for Computational Linguistics.

\bibitem[{Srinivasan(2017)}]{srinivasan2017reimagining}
Ramesh Srinivasan. 2017.
\newblock \href {https://nyupress.org/9781479862962/whose-global-village/}
  {\emph{Whose Global Village? {R}ethinking How Technology Shapes Our World}}.
\newblock New York University Press, New York.

\bibitem[{Trichopoulos(2023)}]{trichopoulos2023large}
Georgios Trichopoulos. 2023.
\newblock \href {https://doi.org/10.1145/3609987.3610018} {Large language
  models for cultural heritage}.
\newblock In \emph{Proceedings of the 2nd International Conference of the ACM
  Greek SIGCHI Chapter}, CHIGREECE '23, New York, NY, USA. Association for
  Computing Machinery.

\bibitem[{van Hooland and Verborgh(2015)}]{vanhooland2014linked}
Seth van Hooland and Ruben Verborgh. 2015.
\newblock \emph{Linked Data for Libraries, Archives and Museums: How to clean,
  Link and Publish your Metadata}.
\newblock Facet.

\bibitem[{Wang(2019)}]{wang-2019-revisiting}
Hongmin Wang. 2019.
\newblock \href {https://doi.org/10.18653/v1/W19-8639} {Revisiting challenges
  in data-to-text generation with fact grounding}.
\newblock In \emph{Proceedings of the 12th International Conference on Natural
  Language Generation}, pages 311--322, Tokyo, Japan. Association for
  Computational Linguistics.

\bibitem[{Wiseman et~al.(2017)Wiseman, Shieber, and
  Rush}]{wiseman-etal-2017-challenges}
Sam Wiseman, Stuart Shieber, and Alexander Rush. 2017.
\newblock \href {https://doi.org/10.18653/v1/D17-1239} {Challenges in
  data-to-document generation}.
\newblock In \emph{Proceedings of the 2017 Conference on Empirical Methods in
  Natural Language Processing}, pages 2253--2263, Copenhagen, Denmark.
  Association for Computational Linguistics.

\bibitem[{Zhang et~al.(2020)Zhang, Kishore, Wu, Weinberger, and
  Artzi}]{zhang2020bertscore}
Tianyi Zhang, Varsha Kishore, Felix Wu, Kilian~Q. Weinberger, and Yoav Artzi.
  2020.
\newblock \href {https://openreview.net/forum?id=SkeHuCVFDr} {{BERTScore}:
  Evaluating text generation with {BERT}}.
\newblock In \emph{8th International Conference on Learning Representations,
  {ICLR} 2020, Addis Ababa, Ethiopia, April 26-30, 2020}. OpenReview.net.

\end{thebibliography}
